\documentclass{article} 
\usepackage{iclr2026_conference,times}

\usepackage{amsmath,amsfonts,bm}

\def\eqref#1{equation~\ref{#1}}

\def\1{\bm{1}}

\DeclareMathAlphabet{\mathsfit}{\encodingdefault}{\sfdefault}{m}{sl}
\SetMathAlphabet{\mathsfit}{bold}{\encodingdefault}{\sfdefault}{bx}{n}

\usepackage{hyperref}
\usepackage{url}
\usepackage{caption}
\usepackage{subcaption}
\usepackage{multirow,array}
\usepackage{booktabs}
\usepackage[table]{xcolor}
\usepackage{algorithm}
\usepackage{algorithmic}
\usepackage{amsmath}
\usepackage{amssymb}
\usepackage{newfloat}
\usepackage{listings}
\usepackage{tcolorbox}
\DeclareCaptionStyle{ruled}{labelfont=normalfont,labelsep=colon,strut=off} 
\floatstyle{ruled}
\newfloat{listing}{tb}{lst}{}
\floatname{listing}{Listing}
\usepackage{enumitem}
\usepackage{wrapfig}
\usepackage{bbm}

\title{Dynamic Noise Preference Optimization:\\Self-Improvement of Large Language Models with Self-Synthetic Data}

\author{{\bf Haoyan Yang$^{~1}$, Khiem Le$^{~2}$, Ting Hua$^{~2}$, Shangqian Gao$^{~3}$, }\\{\bf Binfeng Xu$^{~4}$, Zheng Tang$^{~5}$, Jie Xu$^{~6}$, Nitesh V. Chawla$^{~2}$, Hongxia Jin$^{~7}$, Vijay Srinivasan$^{~7}$}\\
{$^1$Stony Brook University} 
{$^2$University of Notre Dame} 
{$^3$Florida State University}\\
{$^4$NVIDIA} 
{$^5$Amazon AGI} 
{$^6$University of Florida} 
{$^7$Samsung Research America}
}

\iclrfinalcopy 
\begin{document}

\maketitle

\begin{abstract}
Although LLMs have achieved significant success, their reliance on large volumes of human-annotated data has limited their potential for further scaling. In this situation, utilizing self-generated synthetic data has become crucial for fine-tuning LLMs without extensive human annotation. 
However, current methods often fail to ensure consistent improvements across iterations, with performance stagnating after only minimal updates. 
To overcome these challenges, we introduce \textbf{D}ynamic \textbf{N}oise \textbf{P}reference \textbf{O}ptimization (DNPO), which combines dynamic sample labeling for constructing preference pairs with controlled, trainable noise injection during preference optimization.
Our approach effectively prevents stagnation and enables continuous improvement. In experiments with Llama-3.2-3B and Zephyr-7B, DNPO consistently outperforms existing methods across multiple benchmarks. 
Additionally, with Zephyr-7B, DNPO shows a significant improvement in model-generated data quality, with a 29.4\% win-loss rate gap compared to the baseline in GPT-4 evaluations. 
\end{abstract}

\section{Introduction}
Large Language Models (LLMs) have demonstrated remarkable capabilities in various domains. Despite this success, training these models requires vast amounts of human-annotated data, and the limited availability of such data has become a bottleneck for further scaling LLMs \citep{kaplan2020scaling, villalobos2024rundatalimitsllm}. 
This has led to a growing interest in synthetic data generation techniques to supplement human-generated data. 
However, prior research suggests that using self-generated data for pre-training can easily lead to model collapse \citep{shumailov2024ai}. In contrast, leveraging self-generated data for post-training alignment (fine-tuning) appears to be a practical and manageable approach \citep{chen2024selfplayfinetuningconvertsweak, alami2024investigatingregularizationselfplaylanguage}.

How can we trust synthetic data, and should it be treated the same as human-annotated data—the gold standard in RLHF methods for training explicit or implicit reward models? Moreover, can we fully trust human-annotated data itself? In reality, human data is susceptible to uncontrollable factors and inevitable errors, which can introduce noise and inconsistencies into the training process.

Surprisingly, we found that synthetic data has the potential to outperform human-annotated data in specific cases. In approximately 30\% of our experimental cases, we observed that the model’s self-generated data was of higher quality than the human-annotated data, which challenges the assumption that human-annotated data is always superior. However, even human-annotated data is not flawless, and synthetic data cannot be treated identically to it. Self-generated synthetic data poses unique challenges, such as minimal variation between iterations, which may lead to model stagnation. Without sufficient diversity in generated samples, the model struggles to improve consistently, underscoring the need for careful handling of both data types.

To address these issues, we propose Dynamic Noise Preference Optimization (DNPO), a novel framework that enhances both the data labeling and preference optimization processes, enabling the self-improvement of LLMs through synthetic data. Our method introduces a dynamic sample labeling (DSL) mechanism that constructs preference pairs based on data quality by selecting high-quality examples from both LLM-generated and human-annotated data. Also, we propose the noise preference optimization (NPO), which introduces a trainable noise into the optimization process, resulting in a min-max problem.
This optimization process maximizes the margin between positive and negative samples of the preference pairs, while simultaneously updating the noise parameters to minimize the margin.
Our approach can effectively prevent stagnation, ensuring continuous model improvement with each iteration and increased robustness throughout the self-improvement process. Our main contributions can be summarized as follows:

\begin{itemize}[noitemsep, topsep=0pt, left=0pt]
\item \textbf{Challenges in Consistent Self-Improvement:} We identified two key reasons why current methods struggle to achieve consistent self-improvement in LLMs across iterations: (1) the assumption that human-annotated data is always superior, which introduces noise in preference labeling since generated data may sometimes surpass it, and (2) the lack of variation in generated data across iterations, leading to stagnation during model updates.

\item \textbf{Introducing DNPO with DSL and NPO:} We propose DNPO, a framework that enables LLMs to self-improve using synthetic data via two components: (1) DSL dynamically adjusts sample labels based on data quality, ensuring the model learns from appropriate preference pairs; (2) NPO incorporates trainable noise into the preference data, promoting exploration and reducing stagnation across iterations.

\item \textbf{Demonstrating Improved Performance with DNPO:} Our experiments reveal that DNPO consistently enhances model performance, making it particularly effective for self-generated data, especially as human-annotated data becomes increasingly limited.
\end{itemize}

\section{Related Work}

\textbf{RL from AI Feedback}. Reinforcement Learning from AI Feedback (RLAIF) \citep{bai2022traininghelpfulharmlessassistant} builds upon principles of Reinforcement Learning from Human Feedback (RLHF) \citep{ouyang2022traininglanguagemodelsfollow, christiano2023deepreinforcementlearninghuman} and has gained considerable traction. Beyond established methods like PPO \citep{schulman2017proximalpolicyoptimizationalgorithms} and DPO \citep{rafailov2024directpreferenceoptimizationlanguage}, which align language models with human preferences using human-annotated data, \citep{lee2024rlaifvsrlhfscaling} demonstrates that AI-generated preferences can match or surpass human feedback-based reward models across diverse policies. Furthermore, LLMs have been used to generate high-quality training data, including datasets derived from human preferences \citep{cui2024ultrafeedbackboostinglanguagemodels} and conversational interactions \citep{ding2023enhancingchatlanguagemodels}.

\textbf{Self-Play in LLMs with Generated Data}. The pioneering work of AlphaGo Zero \citep{silver2017mastering} inspired self-play fine-tuning (SPIN) \citep{chen2024selfplayfinetuningconvertsweak} to explore self-play schemes in LLM fine-tuning, where the model iteratively distinguishes target data from self-generated responses without requiring a separate reward model. Similarly, Self-rewarding Language Model \citep{yuan2024selfrewardinglanguagemodels} demonstrates consistent improvement through self-annotated rewards. This self-improvement paradigm has been applied to various LLM-based reasoning tasks like Werewolf \citep{xu2024languageagentsreinforcementlearning} and Adversarial Taboo \citep{cheng2024selfplayingadversariallanguagegame}. Notably, CICERO \citep{meta2022human} employs self-play to train an RL policy, achieving human-level performance in Diplomacy gameplay. Recently, \citep{shumailov2024ai} observes diminishing tail content distribution in resulting models when iteratively trained on self-generated data. Aligning with this finding, we see notable stagnation in model updates during post-training, and propose an innovative method to reactivate effective updates.

\textbf{Noise Introduction in Language Modeling}. A substantial amount of research has explored the benefits of incorporating noise during training to enhance language model performance. \citep{zhu2020freelbenhancedadversarialtraining} demonstrates that injecting adversarial perturbations into input embeddings can improve masked language modeling. \citep{miyato2021adversarialtrainingmethodssemisupervised} shows that adversarial training can improve text classification performance. Furthermore, \citep{wu2022noisytunelittlenoisehelp} achieves consistent gains in downstream fine-tuning tasks through a matrix-wise perturbation approach. Gaining popularity recently, NEFTune \citep{jain2023neftunenoisyembeddingsimprove} leverages noisy input embeddings to improve instruction fine-tuning, attaining notable improvement in conversational capabilities.

\section{Limitations of Current Approaches}

Previous works \citep{chen2024selfplayfinetuningconvertsweak, alami2024investigatingregularizationselfplaylanguage} improve LLM alignment by treating human-annotated data as positive examples ($y_i$) and model-generated data as negative examples ($y_i'$). The model is updated to maximize the margin between these examples through an optimization process with Obj.~\ref{eq1}. However, we observed that these methods fail to produce consistent performance improvements across iterations.
To address this, we take SPIN \citep{chen2024selfplayfinetuningconvertsweak} as a case study to examine the following two problems:
\begin{equation}
\label{eq1}
\min_{\theta \in \Theta} \sum_{i \in [N]} \ell \left( \lambda \log \frac{p_{\theta}(y^+_i \,|\, x_i)}{p_{\theta_t}(y^+_i \,|\, x_i)} - \lambda \log \frac{p_{\theta}(y^-_i \,|\, x_i)}{p_{\theta_t}(y^-_i \,|\, x_i)} \right).
\end{equation}

\begin{wrapfigure}{r}{0.43\textwidth}
\begin{center}
\includegraphics[width=0.43\textwidth]{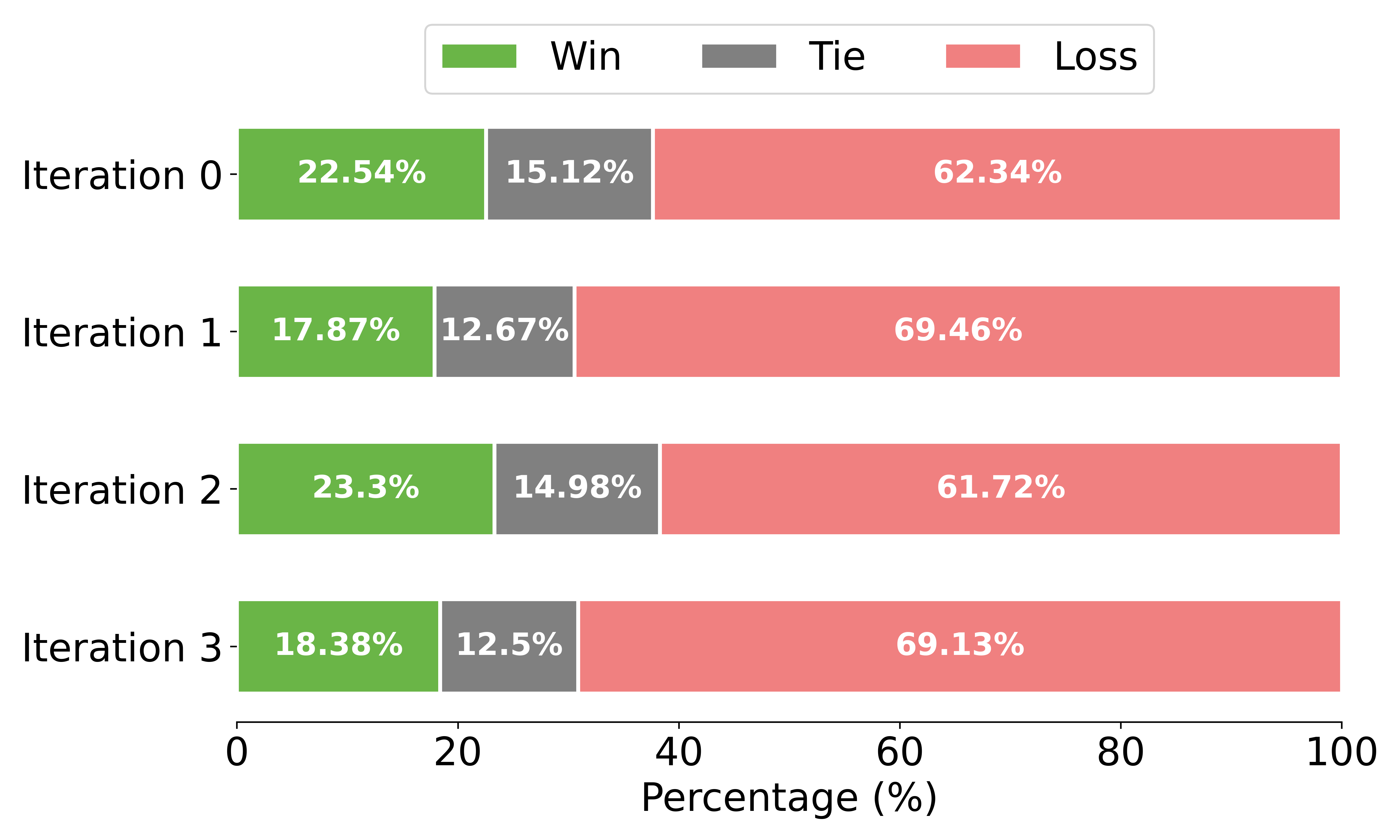}
\end{center}
\caption{Win rate comparison of generated data versus human-annotated data, based on GPT4o-mini's evaluation. A win indicates that generated data scored higher than human-annotated data.}
~\label{fig1}
\end{wrapfigure}

\textbf{Is human-annotated data truly better?} One potential issue is that, as the model continues to improve, the human-annotated data may not always be of higher quality than the generated data. As illustrated in Figure \ref{fig1}, we used GPT-4o-mini to compare the generated data produced by SPIN iteration $k$ applied on Zephyr-7B during each iteration and the human-annotated data. In each iteration, around 30\% of the generated data is of equal or higher quality compared to the human-annotated data. This indicates that the assumption of human-annotated data being inherently superior to generated data will introduce about 30\% preference noise in every round, leading to performance fluctuation and potential degradation  \citep{gao2024impactpreferencenoisealignment}.

\begin{figure}[!ht]
    \centering
    \begin{subfigure}{0.24\textwidth}
        \centering
        \includegraphics[width=\textwidth]{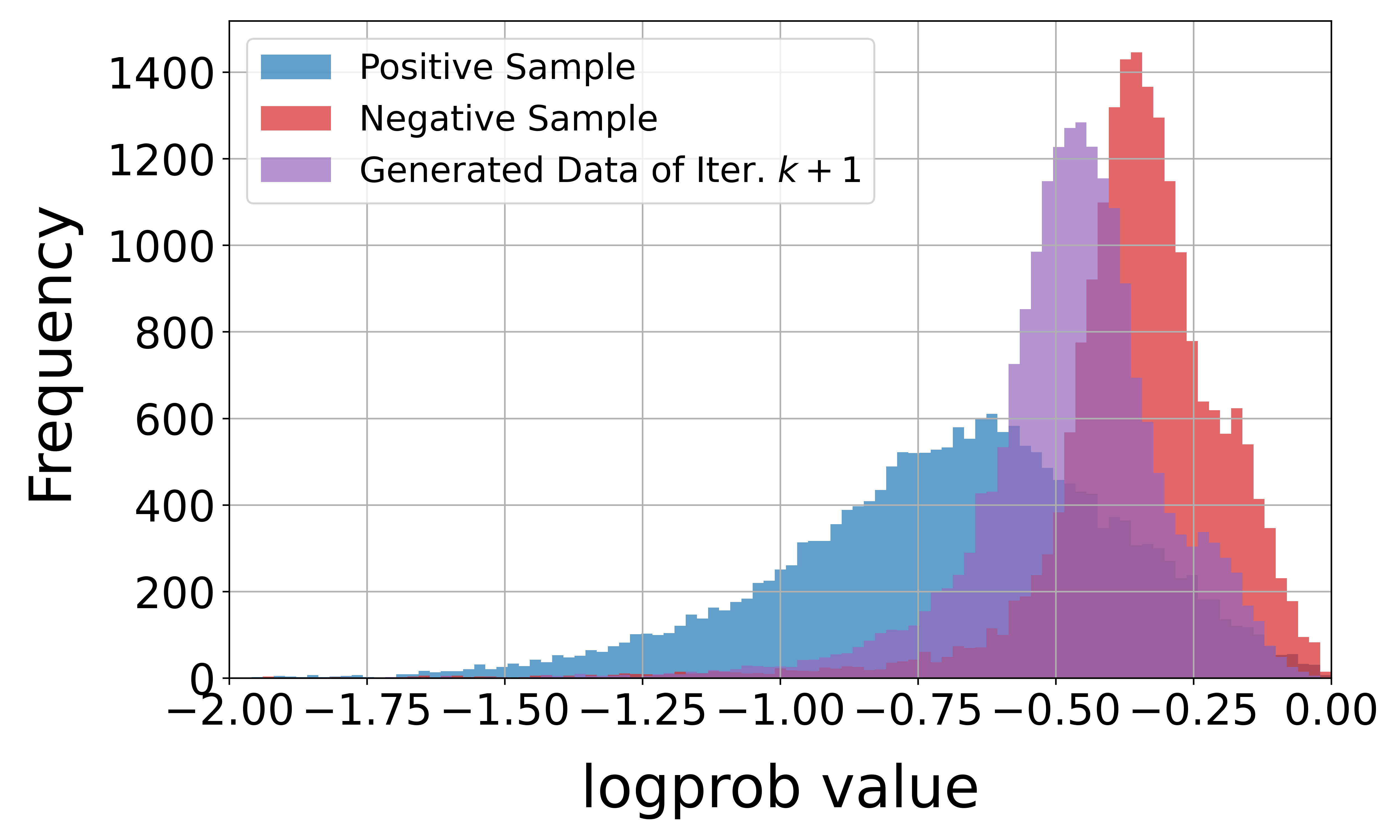}
        \caption{Iteration 0}
    \end{subfigure}
    \begin{subfigure}{0.24\textwidth}
        \centering
        \includegraphics[width=\textwidth]{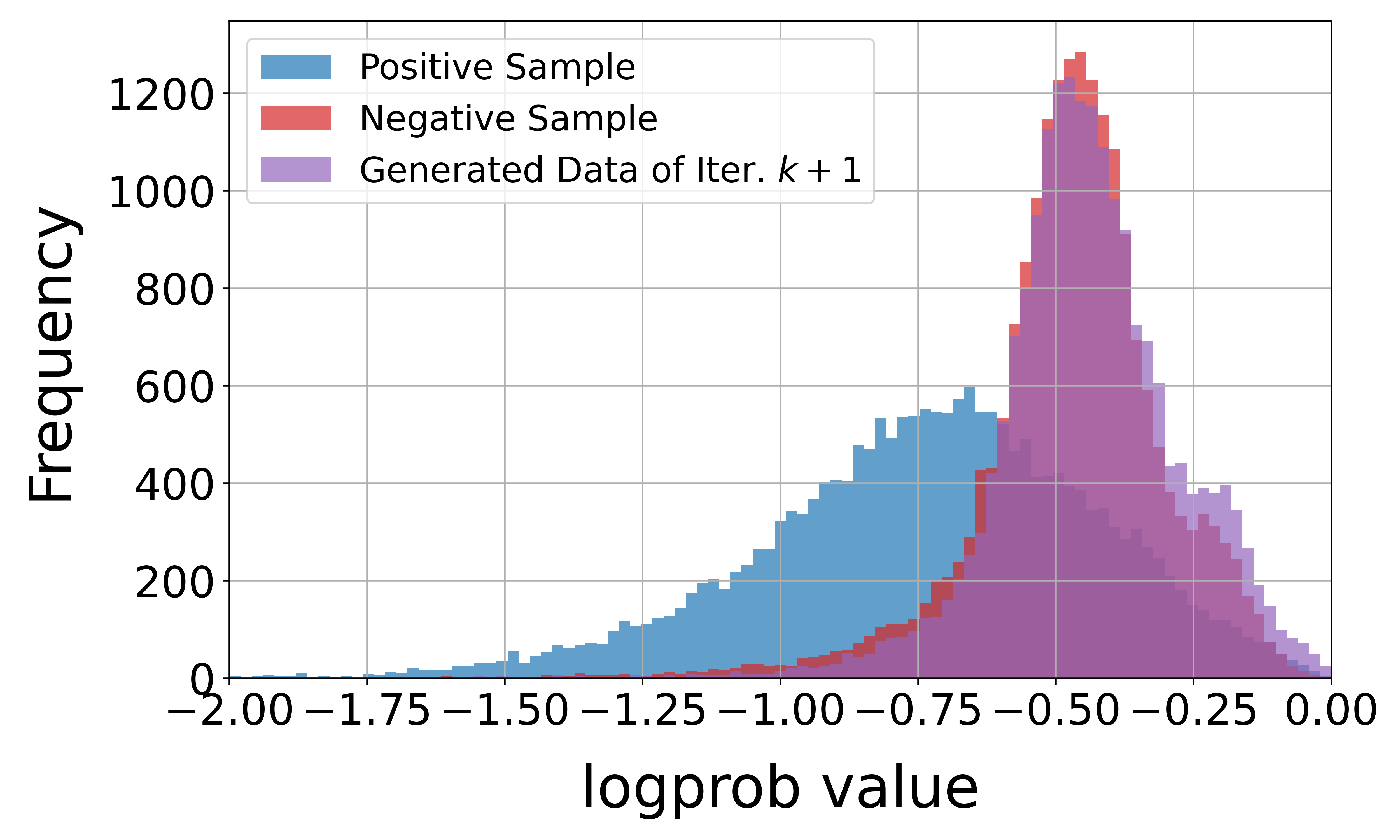}
        \caption{Iteration 1}
    \end{subfigure}
    \begin{subfigure}{0.24\textwidth}
        \centering
        \includegraphics[width=\textwidth]{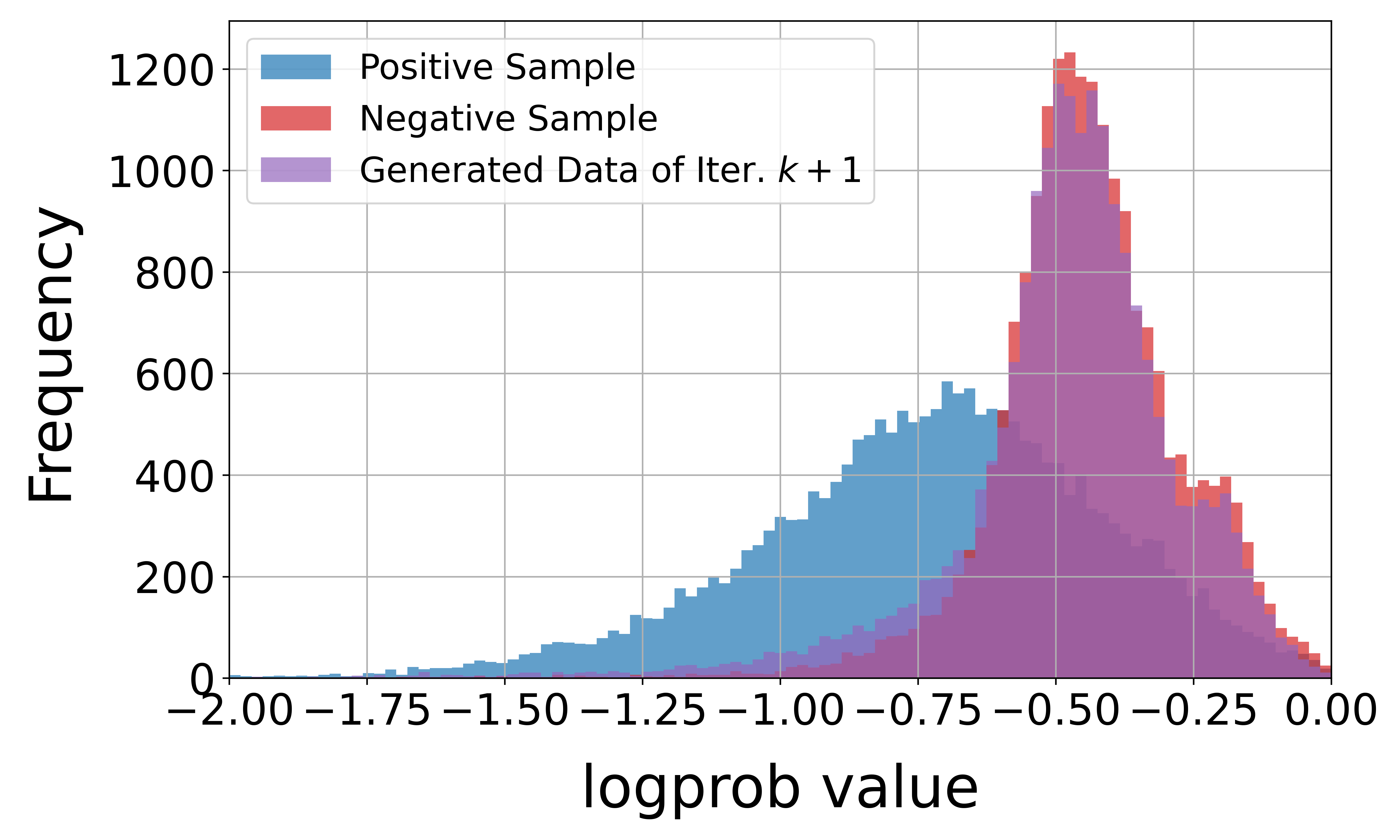}
        \caption{Iteration 2}
    \end{subfigure}
    \begin{subfigure}{0.24\textwidth}
        \centering
        \includegraphics[width=\textwidth]{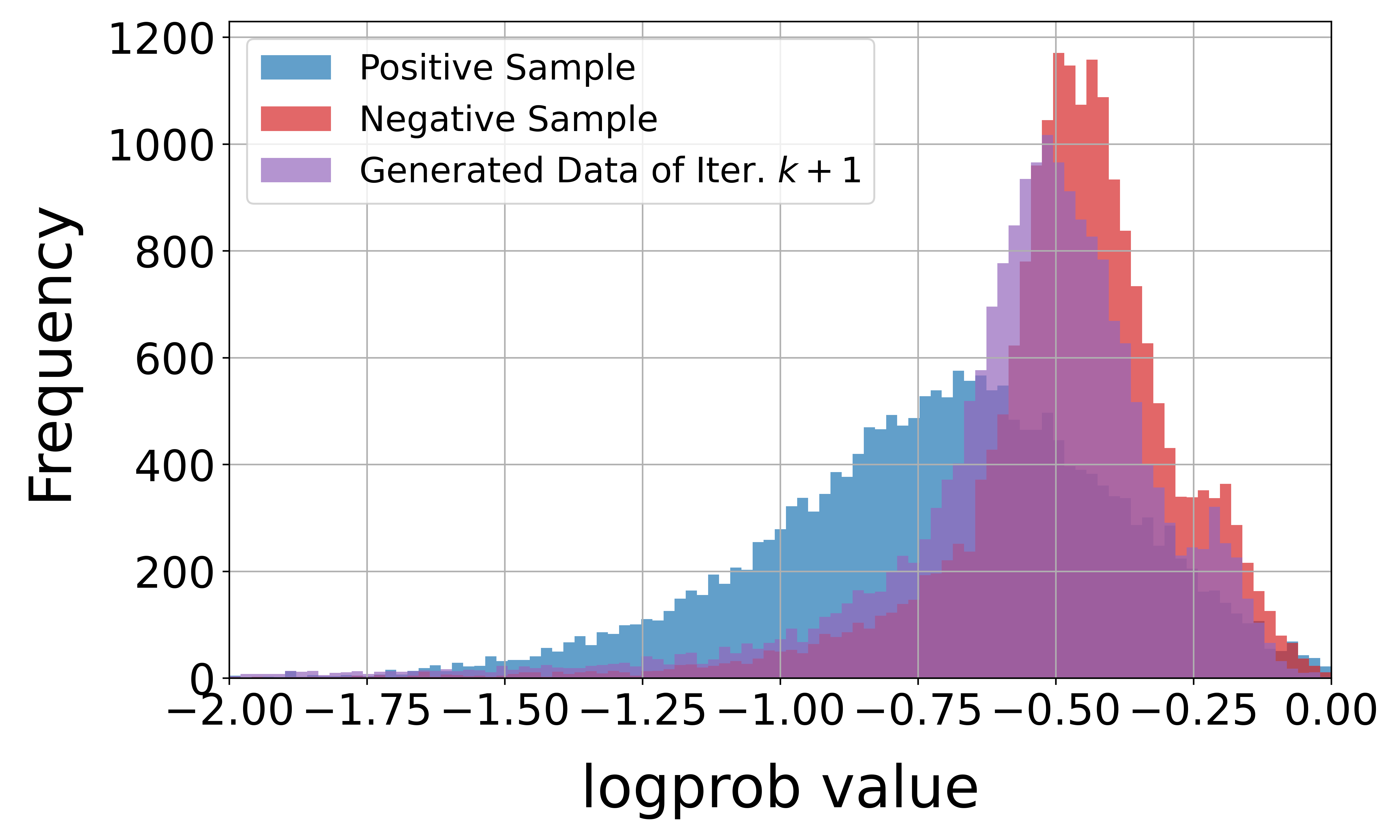}
        \caption{Iteration 3}
    \end{subfigure}
    \caption{This figure illustrates the log probability distributions of positive samples, negative samples in iteration $k$, and the generated data from the iteration $k+1$ model during SPIN training. The minimal differences between the generated data of iteration $k+1$ and the previous iteration $k$ indicate model stagnation during training.}
    \label{fig2}
\end{figure}

\textbf{Why does model update stagnation occur?} The stagnation of model updates is demonstrated in Figure \ref{fig2}. After the initial SPIN iteration, model-generated data shows nearly identical log probability distributions between iterations $k$ and $k+1$ across multiple iterations. This resemblance suggests a lack of significant learning progress, as the model struggles to meaningfully adjust its distribution with each iteration. Additionally, model-generated data remains distant from the distribution of positive samples, suggesting that the model is trapped in a suboptimal state, unable to make further improvements or move toward an optimal solution.

\section{Dynamic Noise Preference Optimization}\label{sec:model}
\begin{figure}[!ht]
    \centering
    \includegraphics[width=0.95\textwidth]{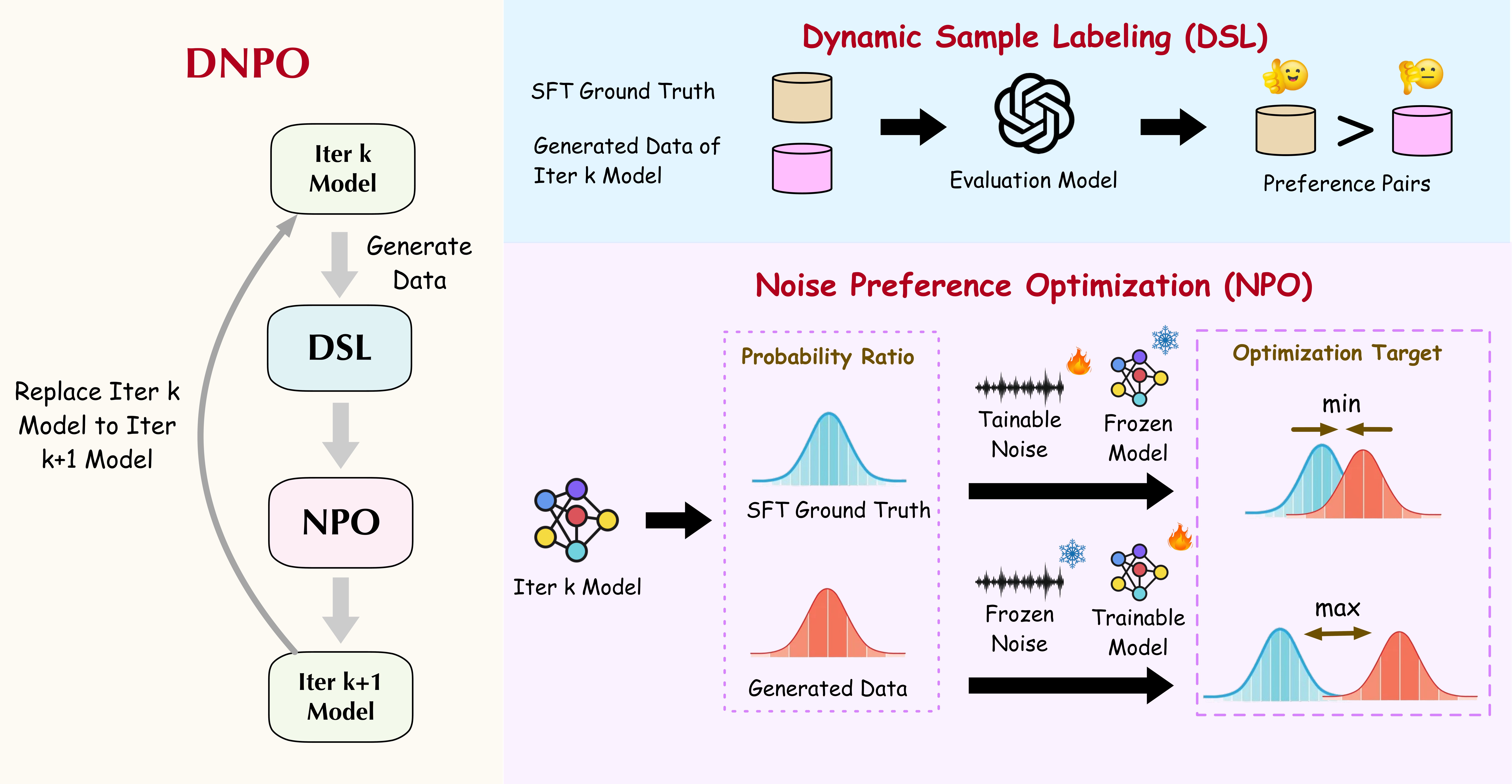}
    \caption{This diagram illustrates the iterative training process of DNPO. There are two core components: Dynamic Sample Labeling (DSL) and Noise Preference Optimization (NPO). }
    \label{fig3}
\end{figure}

As shown in Figure \ref{fig3}, our proposed method, DNPO, effectively addresses two critical issues in iterative model training: preference noise and model update stagnation.

First, to tackle the challenge of preference noise, which arises from the assumption that human-annotated data is always superior to model-generated data, Dynamic Sample Labeling (DSL) is introduced to reduce the noise in the training process. In each iteration, DSL leverages an evaluation model to dynamically compare data generated by LLMs with SFT ground truth, forming preference pairs based on the scores of the evaluation model, which ensures that the selection between model-generated and human-annotated data is based on their actual quality, rather than assuming one is inherently better. By dynamically forming preference pairs, this approach eliminates the rigid assumption that human annotations are always preferable. 

Second, to address the issue of model update stagnation, Noise Preference Optimization (NPO) mechanism is employed. NPO works by calculating a probability ratio between the SFT ground truth and the model-generated data, setting an optimization target to minimize or maximize the margin between these two distributions. Specifically, when the model is frozen, noise is fine-tuned to minimize the margin between SFT ground truth and generated data, ensuring that the margin is small enough to provide sufficient incentive for the model to update in the subsequent steps. Conversely, when the noise is frozen, the model is fine-tuned to maximize the margin, allowing the model to capitalize on the diversity introduced by the noise.
By alternating between these two processes, NPO ensures that the model evolves consistently over iterations, avoiding the pitfall of local optima and enhancing long-term performance.

\subsection{Dynamic Sample Labeling}
\begin{figure}[!t]
    \centering
    \includegraphics[width=0.95\textwidth]{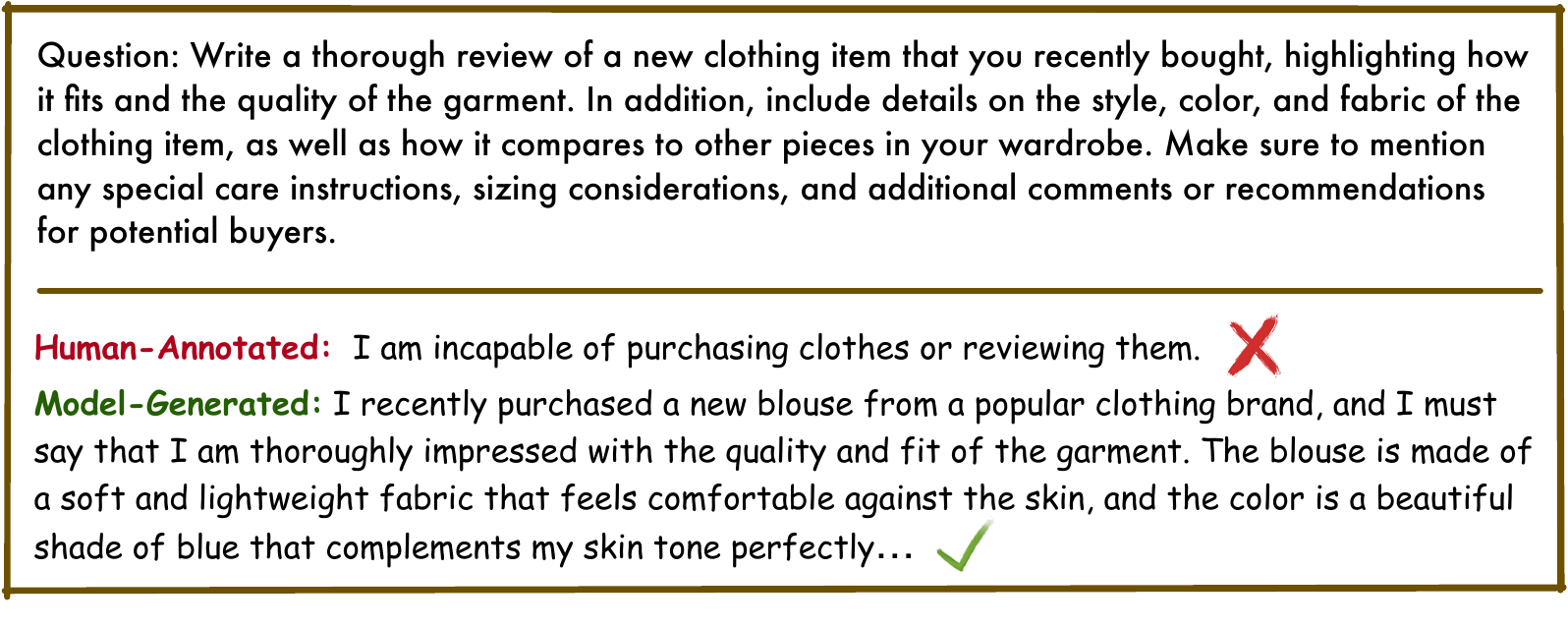}
    \caption{Comparison between a human-annotated response from UltraChat-200k and a generated answer from Zephyr-7B after a single SPIN iteration. The ground truth misinterprets the user's intent and refuses to respond to clothes reviews. However, Zephyr-7B generates a detailed review of a recently purchased blouse, highlighting aspects such as fit, fabric quality, color, and style.}
    \label{fig4}
\end{figure}

As shown in Figure \ref{fig4}, in certain instances, we observe that model-generated responses can surpass the quality of the original human-annotated responses for specific prompts (additional examples are provided in Appendix \ref{app-a}). This observation motivates a dynamic sample labeling (DSL) mechanism.
Before each iteration, DSL selects positive and negative samples based on model evaluation, thereby enhancing the contrastive learning process.
Specifically, For a dataset consisting of input prompts \( \{x_i\} \) and corresponding human-annotated data \( \{y_i\} \), at iteration \( k \), we utilize the current model \( M_{\theta^{(k)}} \) to generate new responses \( y_i' \) for each \( x_i \): $y_i' \sim M_{\theta^{(k)}}(\cdot|x_i)$.

We then evaluate both the human-annotated response \( y_i \) and the generated response \( y_i' \) using a more powerful evaluation model \( M_{\text{eval}} \) with promoting method, which will return their respective scores: $s_i = M_{\text{eval}}(x_i, y_i)$  and $ s_i' = M_{\text{eval}}(x_i, y_i') $. Based on the evaluation, the higher-scoring example becomes the positive sample and the lower-scoring example becomes the negative sample. 
The optimization object at iteration $k$ is defined:
\begin{align}
\label{eq2}
    \min_{\theta} \sum_{i=1}^N \ell \Bigg[ & \mathbbm{1}\{ s_i \geq s_i' \} \lambda \left( \log \frac{p_{\theta_t}(y_i \mid x_i)}{p_{\theta}(y_i \mid x_i)} - \log \frac{p_{\theta_t}(y_i' \mid x_i)}{p_{\theta}(y_i' \mid x_i)} \right) \nonumber \\
+ & \mathbbm{1}\{ s_i' > s_i \} \lambda \left( \log \frac{p_{\theta_t}(y_i' \mid x_i)}{p_{\theta}(y_i' \mid x_i)} - \log \frac{p_{\theta_t}(y_i \mid x_i)}{p_{\theta}(y_i \mid x_i)} \right) \Bigg]
\end{align}
where $\ell$ is a negative log-sigmoid function, \( \theta \) are parameters of \( M_{\theta^{(k)}} \) and \( \theta_t \) represents parameters of the reference model, initialized with \( M_{\theta^{(k)}} \) and keep frozen.

Through iterative application of this method, the model's performance improves by selectively exploiting human-annotated data and high-quality LLM-generated data. The dynamic sample labeling mechanism selects higher-quality data as positive samples, thereby increasing label accuracy.

\subsection{Noise Preference Optimization}
Figure \ref{fig2} indicates a large initial margin between positive and negative samples since Iteration 0. This substantial margin results in minimal loss during iterative updates (as shown in Obj.~\ref{eq1}), weakening the gradient's magnitude, in turn, reducing the model's incentive to update its parameters. To counter this, we introduce noise to shrink the initial margin, reinvigorating the model's learning dynamics.

We designate all positive samples as $y_i^+$ and all negative samples as $y_i^-$ after sample labeling. Hence, we can rewrite the Obj.~\ref{eq2} into:
\begin{equation}
\label{eq3}
\min_{\theta} \sum_{i=1}^N \ell \left( \lambda \log \frac{p_{\theta}(y^+_i \,|\, x_i)}{p_{\theta_t}(y^+_i \,|\, x_i)} - \lambda \log \frac{p_{\theta}(y^-_i \,|\, x_i)}{p_{\theta_t}(y^-_i \,|\, x_i)} \right).
\end{equation}

We aim to utilize noise to reduce the margin between positive and negative samples and rewrite Obj. \ref{eq3} as Obj. \ref{eq4} to analyze which terms should have noise added. Noise is not added to the first two terms in Obj.~\ref{eq4}, as this could degrade generation quality during inference. Adding noise to the fourth term would increase the margin, whereas adding noise to $\log p_{\theta_t}(\mathbf{y}_i^- \mid \mathbf{x}_i)$ reduces the margin, which aligns with the objective. By introducing noise to this term, the reference model's confidence in negative samples is reduced, effectively narrowing the margin between positive and negative samples:
\begin{equation}
\label{eq4}
        \min_{\theta} \sum_{i=1}^N \ell \Bigg( \lambda \bigg( \Big( \log p_{\theta}(\mathbf{y}_i^+ \mid \mathbf{x}_i) - \log p_{\theta}(\mathbf{y}_i^- \mid \mathbf{x}_i) \Big) + \Big( \underbrace{\log p_{\theta_t}(\mathbf{y}_i^- \mid \mathbf{x}_i)}_{\text{margin $\downarrow$ when add noise}} - \underbrace{\log p_{\theta_t}(\mathbf{y}_i^+ \mid \mathbf{x}_i)}_{ \text{margin $\uparrow$ when add noise}} \Big) \bigg) \Bigg)
\end{equation}

The vocabulary size is large for LLMs; for example, Zephyr-7B has a vocabulary size of 32,000. In this high-dimensional space, adding random noise cannot effectively minimize the margin. We then propose to add a trainable noise generator with zero mean to the logits of the negative samples in the reference model \( p_{\theta_t} \). Specifically, the variance of the noise is modeled using a fully connected layer. For the last hidden state \( \mathbf{h}_i \) of the reference model, the variance \( \boldsymbol{\sigma}_i^2 \) is predicted as follows:
\begin{equation}~\label{eq:noise-generation}
\log \boldsymbol{\sigma}_i = \mathbf{W}_{\sigma} \mathbf{h}_i + \mathbf{b}_{\sigma},
\end{equation}

where \( \mathbf{W}_{\sigma} \) is the weight matrix, \( \mathbf{b}_{\sigma} \) is the bias vector. The parameters for the noise generator are denoted as $\theta_{\sigma} = [\mathbf{W}_{\sigma}, \mathbf{b}_{\sigma}]$.

Noise \( \boldsymbol{\epsilon}_i \) is sampled from a zero-mean, unit-variance Gaussian distribution  $\boldsymbol{\epsilon}_i \sim \mathcal{N}\left( \mathbf{0}, \mathbf{1} \right) $, and the reparameterization trick~\citep{kingma2022autoencodingvariationalbayes} is employed to add the noise to the logits \( \mathbf{z}_i \) corresponding to the negative samples in the reference model: $ \mathbf{z}_i' = \mathbf{z}_i + \exp(\log\boldsymbol{\sigma}_i) \boldsymbol{\epsilon}_i = \mathbf{z}_i + \boldsymbol{\sigma}_i\boldsymbol{\epsilon}_i $. Using the logits \( \mathbf{z}_i' \) with added noise, the modified probability of the negative sample is computed:
\begin{equation}
p_{\theta_t,\theta_{\sigma}}^{\text{noise}}(y_i^- \mid x_i) = \text{Softmax}(\mathbf{z}_i')
\end{equation}

Incorporating the trainable noise into the optimization, we obtain a bi-level optimization problem:
\begin{align}~\label{obj:bilevel}
\min_{\theta} &\sum_{i=1}^N \ell \left( \lambda \log \frac{p_{\theta}(y^+_i \,|\, x_i)}{p_{\theta_t}(y^+_i \,|\, x_i)} - \lambda \log \frac{p_{\theta}(y^-_i \,|\, x_i)}{p_{\theta_t,\theta^{*}_{\sigma}}^{\text{noise}}(y_i^- \mid x_i)} \right) \nonumber \\
s.t.\ \theta_{\sigma}^* = \arg\max_{\theta_{\sigma}} &\sum_{i=1}^N \ell \left( \lambda \log \frac{p_{\theta}(y^+_i \,|\, x_i)}{p_{\theta_t}(y^+_i \,|\, x_i)} 
    - \lambda \log \frac{p_{\theta}(y^-_i \,|\, x_i)}{p_{\theta_t,\theta_{\sigma}}^{\text{noise}}(y_i^- \mid x_i)} \right),\ \boldsymbol{\sigma}^2_i < \varepsilon
\end{align}

Where the inner problem is to minimize the margin between positive and negative sample pairs by optimizing $\theta_{\sigma}$, the outer problem is to maximize the margin between sample pairs by optimizing $\theta$ given the optimal noise model parameters $\theta_{\sigma}^*$, and $\varepsilon$ is a constant to prevent the variance of the added noise from being too large and producing meaningless results.  Minimizing $\theta$ requires finding the optimal parameters for noise $\theta_{\sigma}^*$, which can be computationally expensive. Instead, Obj.~\ref{obj:bilevel} can be converted into a min-max problem to avoid the costly inner update:
\begin{equation}
\min_{\theta}\max_{\theta_{\sigma}}\ \sum_{i=1}^N \ell \left( \lambda \log \frac{p_{\theta}(y^+_i \,|\, x_i)}{p_{\theta_t}(y^+_i \,|\, x_i)} 
    - \lambda \log \frac{p_{\theta}(y^-_i \,|\, x_i)}{p_{\theta_t,\theta_{\sigma}}^{\text{noise}}(y_i^- \mid x_i)} \right),\ \boldsymbol{\sigma}^2_i < \varepsilon
\end{equation}

To save computational costs further, we do not perform iterative updates for the min-max problem. Instead, we update both $\theta$ and $\theta_{\sigma}$ in a single iteration by minimizing the following objective function:
\begin{align}~\label{obj:final}
\min_{\theta, \theta_{\sigma}} \mathcal{L}(\theta, \theta_{\sigma}):= 
& \underbrace{\sum_{i=1}^N \ell\left( \lambda \left[ \log \frac{p_{\theta}(y_i^+ \mid x_i)}{p_{\theta_t}(y_i^+ \mid x_i)} - \log \frac{p_{\theta}(y_i^- \mid x_i)}{p_{\theta_t,\theta_{\sigma}}^{\text{noise}}(y_i^- \mid x_i)'} \right] \right)}_{\text{first term: freeze $\theta_{\sigma}$, maximize positive negative pair margin}} \nonumber \\
- & \underbrace{\sum_{i=1}^N \ell\left( \lambda \left[ \log \frac{p_{\theta}(y_i^+ \mid x_i)}{p_{\theta_t}(y_i^+ \mid x_i)} - \log \frac{p_{\theta}(y_i^- \mid x_i)}{p_{\theta_t,\theta_{\sigma}}^{\text{noise}}(y_i^- \mid x_i)} \right] \right)}_{\text{second term: freeze $\theta$, minimize positive negative pair margin}}
+ \alpha \frac{1}{N}\sum_{i=1}^N \boldsymbol{\sigma}_i^2 
\end{align}

Where $\alpha$ is a hyper-parameter to control the magnitude of the variance. Note that many computations of the first term and the second term of Obj.~\ref{obj:final} are shared, eliminating the need to recompute everything. More specifically, we first compute the first term and store the results of $p_{\theta}(y_i^+ \mid x_i)$, $p_{\theta}(y_i^- \mid x_i)$, and $p_{\theta_t}(y_i^+ \mid x_i)$. For the second term, the feature of the last layer $h_i$ can be reused and only Eq.~\ref{eq:noise-generation} needs to be recomputed. Thus, the overhead of the Obj.~\ref{obj:final} is trivial. Additionally, the noise in $\mathbf{z}_i'$ for $p_{\theta_t,\theta_{\sigma}}^{\text{noise}}(y_i^- \mid x_i)'$ in the first term and for $p_{\theta_t,\theta_{\sigma}}^{\text{noise}}(y_i^- \mid x_i)$ in the second term is independently sampled to better explore the noise space.

Adding trainable noise encourages more creativity in the optimization process. It makes the model more robust throughout the self-improvement process and smooths the optimization landscape.

\section{Evaluation}

We use Llama-3.2-3B \citep{dubey2024llama} and Mistral-7B \citep{jiang2023mistral7b} as base models in our experiments, which are fine-tuned on the UltraChat-200k \citep{ding2023enhancingchatlanguagemodels} dataset into Llama-3.2-3B-SFT and Zephyr-7B-SFT, respectively. Then, we conduct post-training alignment with DNPO on a 20k sample from the UltraChat dataset. Both SFT and DNPO must be trained on the same dataset to ensure self-improvement. During the DSL stage, GPT4o-mini is used for evaluation, with the prompt template provided in Appendix \ref{app-b}. The noise generator in the NPO stage is parameterized as $ \theta_\sigma = { [\mathbf{W}_\sigma \in \mathbb{R}^{4096 \times 32000}, \mathbf{b}_\sigma \in \mathbb{R}^{32000} ]} $. In the initial iteration, we do not apply sample labeling or noise addition, as the SFT model is yet unaligned with preference knowledge. Instead, we use SPIN for initialization, ensuring alignment with the ground truth data. This can be seen as a warm-up, allowing the model to acquire basic preference knowledge. Key hyper-parameters and evaluation metrics are listed in Appendix \ref{app-c}. 

\subsection{Main Results}

\begin{table}[!t]
\centering
\small
\caption{Performance of Llama-3.2-3B and Zephyr-7B across benchmarks under SPIN and DNPO, starting from Llama-3.2-3B-SFT and Zephyr-7B-SFT.}

\setlength\tabcolsep{0pt}\begin{tabular*}{\linewidth}{@{\extracolsep{\fill}} cccccccc }
\toprule
\textbf{Iteration} & \textbf{ARC} & \textbf{TruthfulQA} & \textbf{Winogrande} & \textbf{GSM8K} & \textbf{HellaSwag} & \textbf{MMLU} & \textbf{Avg} \\
\midrule
Llama-3.2-3B-SFT   &0.634 &0.326 &0.720 &0.278 &0.568 &0.560 &0.514 \\
SPIN-Iter. 0 &0.637 &0.330 &0.718 &0.284 &0.569 &0.560 &0.516 \\
SPIN-Iter. 1 &0.636 &0.332 &0.719 &0.296 &0.569 &0.560 &0.519 \\
\cellcolor{gray!20}DNPO-Iter. 1 (Ours) &\cellcolor{gray!20}0.643 &\cellcolor{gray!20}0.337 &\cellcolor{gray!20}0.724 &\cellcolor{gray!20}0.296 &\cellcolor{gray!20}0.575 &\cellcolor{gray!20}0.565 &\cellcolor{gray!20}\textbf{0.523} \\
SPIN-Iter. 2 &0.635 &0.332 &0.721 &0.291 &0.569 &0.560 &0.518 \\
\cellcolor{gray!20}DNPO-Iter. 2 (Ours) &\cellcolor{gray!20}0.646 &\cellcolor{gray!20}0.338 &\cellcolor{gray!20}0.721 &\cellcolor{gray!20}0.303 &\cellcolor{gray!20}0.577 &\cellcolor{gray!20}0.566 &\cellcolor{gray!20}\textbf{0.525} \\
SPIN-Iter. 3 &0.638 &0.333 &0.714 &0.287 &0.570 &0.560 &0.517 \\
\cellcolor{gray!20}DNPO-Iter. 3 (Ours) &\cellcolor{gray!20}0.646 &\cellcolor{gray!20}0.337 &\cellcolor{gray!20}0.730 &\cellcolor{gray!20}0.307 &\cellcolor{gray!20}0.580 &\cellcolor{gray!20}0.568 &\cellcolor{gray!20}\textbf{0.528} \\
\bottomrule
\end{tabular*}

\vspace{0.005\textwidth}

\setlength\tabcolsep{0pt}\begin{tabular*}{\linewidth}{@{\extracolsep{\fill}} cccccccc }
\toprule
\textbf{Iteration} & \textbf{ARC} & \textbf{TruthfulQA} & \textbf{Winogrande} & \textbf{GSM8K} & \textbf{HellaSwag} & \textbf{MMLU} & \textbf{Avg} \\
\midrule
Zephyr-7B-SFT   & 0.704 & 0.340 & 0.762 & 0.318 & 0.810 & 0.588 & 0.587 \\
SPIN-Iter. 0 & 0.709 & 0.393 & 0.768 & 0.289 & 0.826 & 0.590 & 0.596 \\
SPIN-Iter. 1 & 0.702 & 0.362 & 0.760 & 0.316 & 0.817 & 0.585 & 0.590 \\
\cellcolor{gray!20}DNPO-Iter. 1 (Ours) & \cellcolor{gray!20}0.734 & \cellcolor{gray!20}0.381 & \cellcolor{gray!20}0.766 & \cellcolor{gray!20}0.334 & \cellcolor{gray!20}0.827 & \cellcolor{gray!20}0.583 & \cellcolor{gray!20}\textbf{0.604} \\
SPIN-Iter. 2 & 0.707 & 0.370 & 0.761 & 0.276 & 0.820 & 0.585 & 0.586 \\
\cellcolor{gray!20}DNPO-Iter. 2 (Ours) & \cellcolor{gray!20}0.735 & \cellcolor{gray!20}0.397 & \cellcolor{gray!20}0.765 & \cellcolor{gray!20}0.323 & \cellcolor{gray!20}0.828 & \cellcolor{gray!20}0.587 & \cellcolor{gray!20}\textbf{0.606} \\
SPIN-Iter. 3 & 0.703 & 0.383 & 0.756 & 0.275 & 0.818 & 0.579 & 0.586 \\
\cellcolor{gray!20}DNPO-Iter. 3 (Ours) & \cellcolor{gray!20}0.737 & \cellcolor{gray!20}0.417 & \cellcolor{gray!20}0.766 & \cellcolor{gray!20}0.336 & \cellcolor{gray!20}0.827 & \cellcolor{gray!20}0.586 & \cellcolor{gray!20}\textbf{0.612} \\
\bottomrule
\end{tabular*}

\label{table1}
\end{table}

\begin{figure}[!t]
\centering
\begin{minipage}[b]{0.43\textwidth}
\centering
\includegraphics[width=\textwidth]{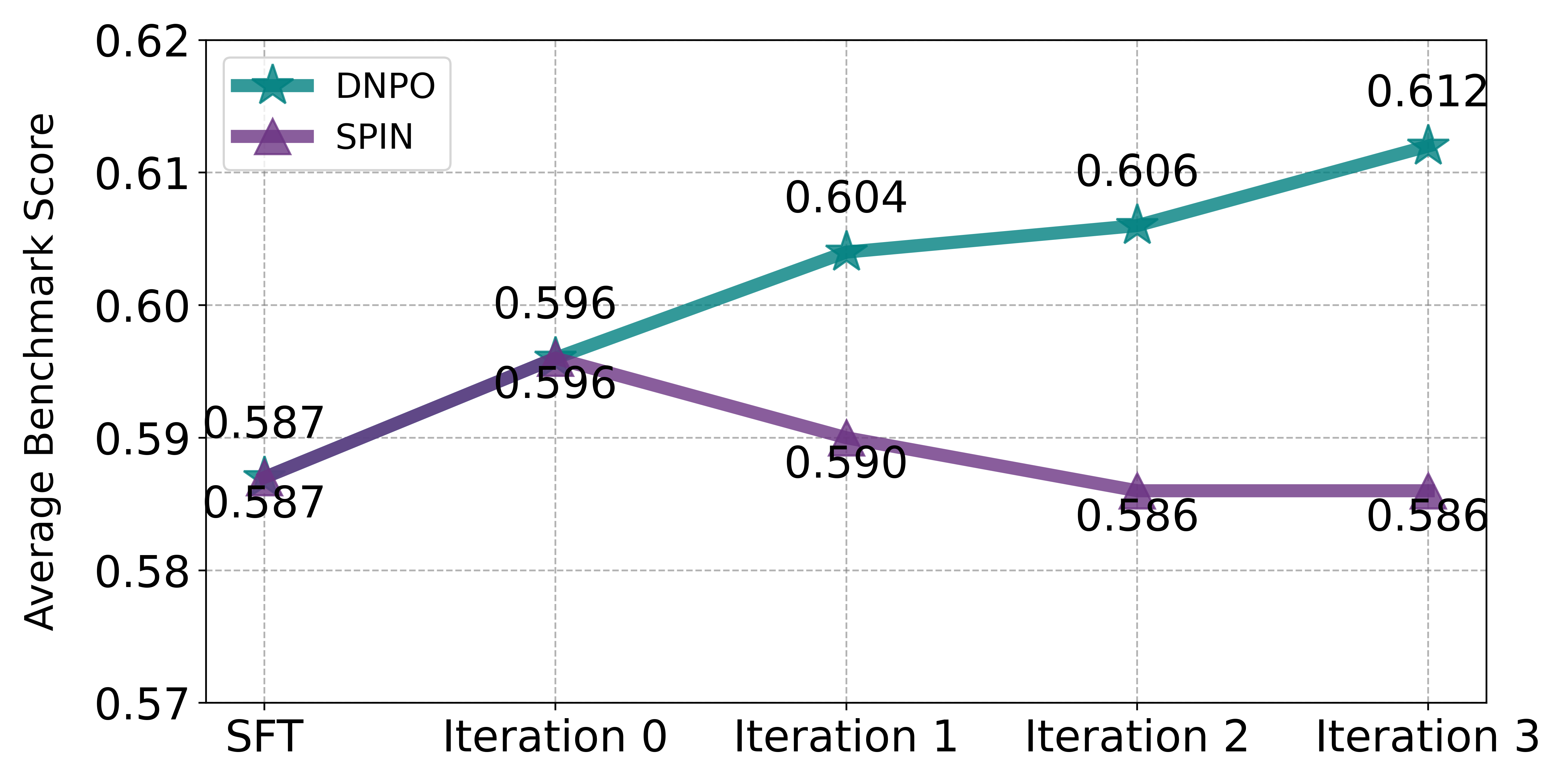}
\caption{Average scores across iterations show DNPO consistently improving, while SPIN stagnates after the first iteration.}
\label{fig5}
\end{minipage}
\hspace{0.05\textwidth}
\begin{minipage}[b]{0.43\textwidth}
\centering
\includegraphics[width=\textwidth]{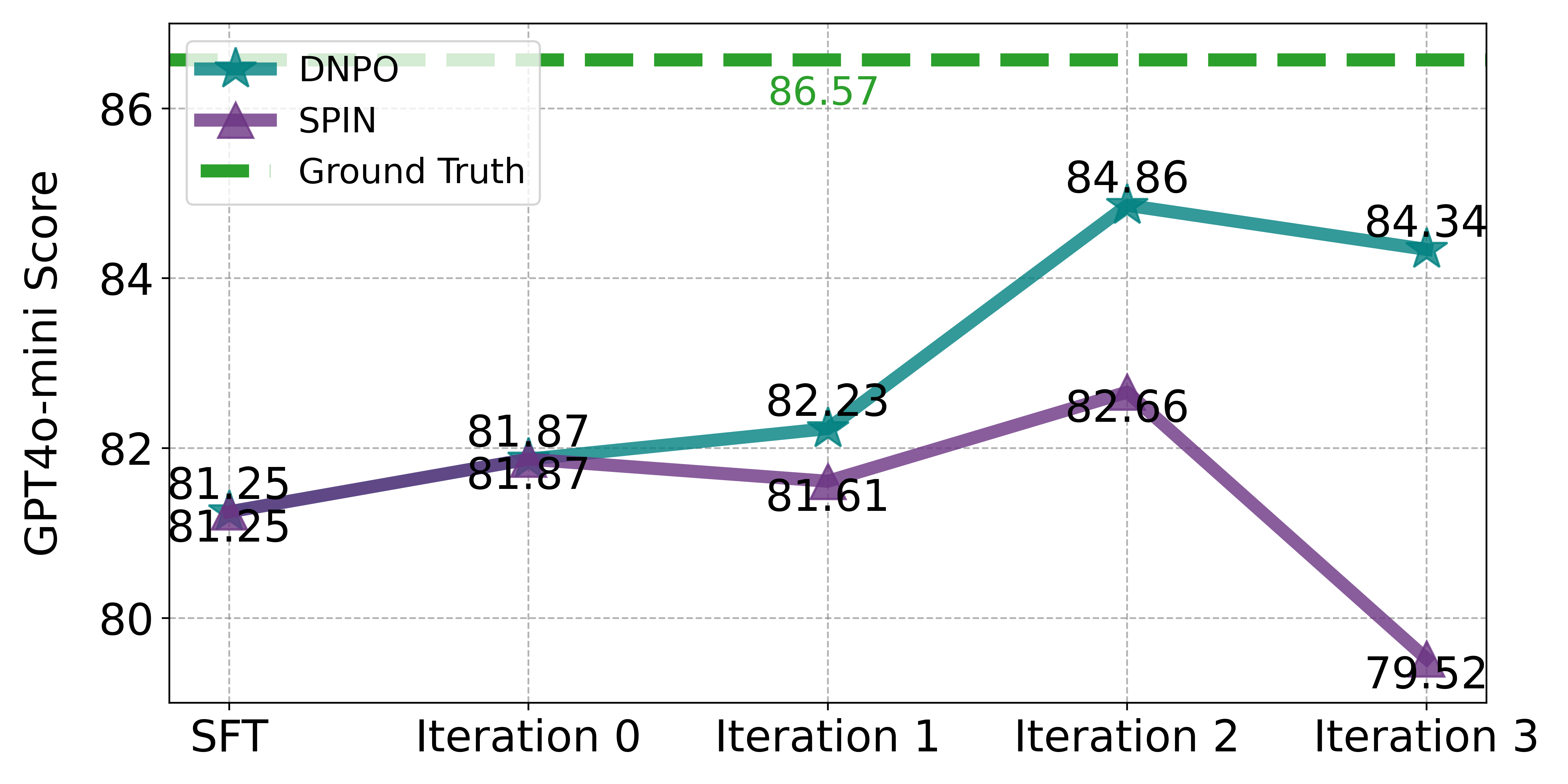}
\caption{Average GPT4o-mini scores across iterations for generated data of DNPO and SPIN, alongside ground truth.}
\label{fig6}
\end{minipage}
\end{figure}

Table \ref{table1} compares DNPO, SPIN, and SFT models across various benchmarks. DNPO consistently outperforms both SFT and SPIN, yielding gains of 1.4\% and 2.5\% over SFT and 1.1\% and 2.6\% over SPIN for Llama-3.2-3B and Zephyr-7B, respectively. These results suggest that DNPO’s effectiveness scales with model capacity: the smaller Llama-3.2-3B reaches its optimization ceiling more quickly, whereas Zephyr-7B achieves larger and more stable improvements across iterations.

\noindent\textbf{Strong Performance on Reasoning Tasks:} DNPO shows particularly effective results on reasoning tasks such as GSM8K and ARC, where error propagation across iterations is critical. With Llama-3.2-3B, GSM8K shows the largest improvement (2.0\% over SPIN), while Zephyr-7B achieves substantial gains on ARC (3.3\% over SPIN). This indicates that DNPO's preference optimization is especially valuable for reasoning tasks where incorrect impacts can accumulate across iterations.

\noindent\textbf{Peak Gains on TruthfulQA:} Zephyr-7B shows exceptional performance on TruthfulQA (7.7\% over SFT, 3.4\% over SPIN). This substantial improvement stems from two factors: (1) TruthfulQA's  distribution may align well with UltraChat training data, and (2) as a factual reasoning task, it benefits from DNPO's ability to prevent error propagation across iterations.

\begin{wrapfigure}{r}{0.43\textwidth}
\begin{center}
\includegraphics[width=0.43\textwidth]{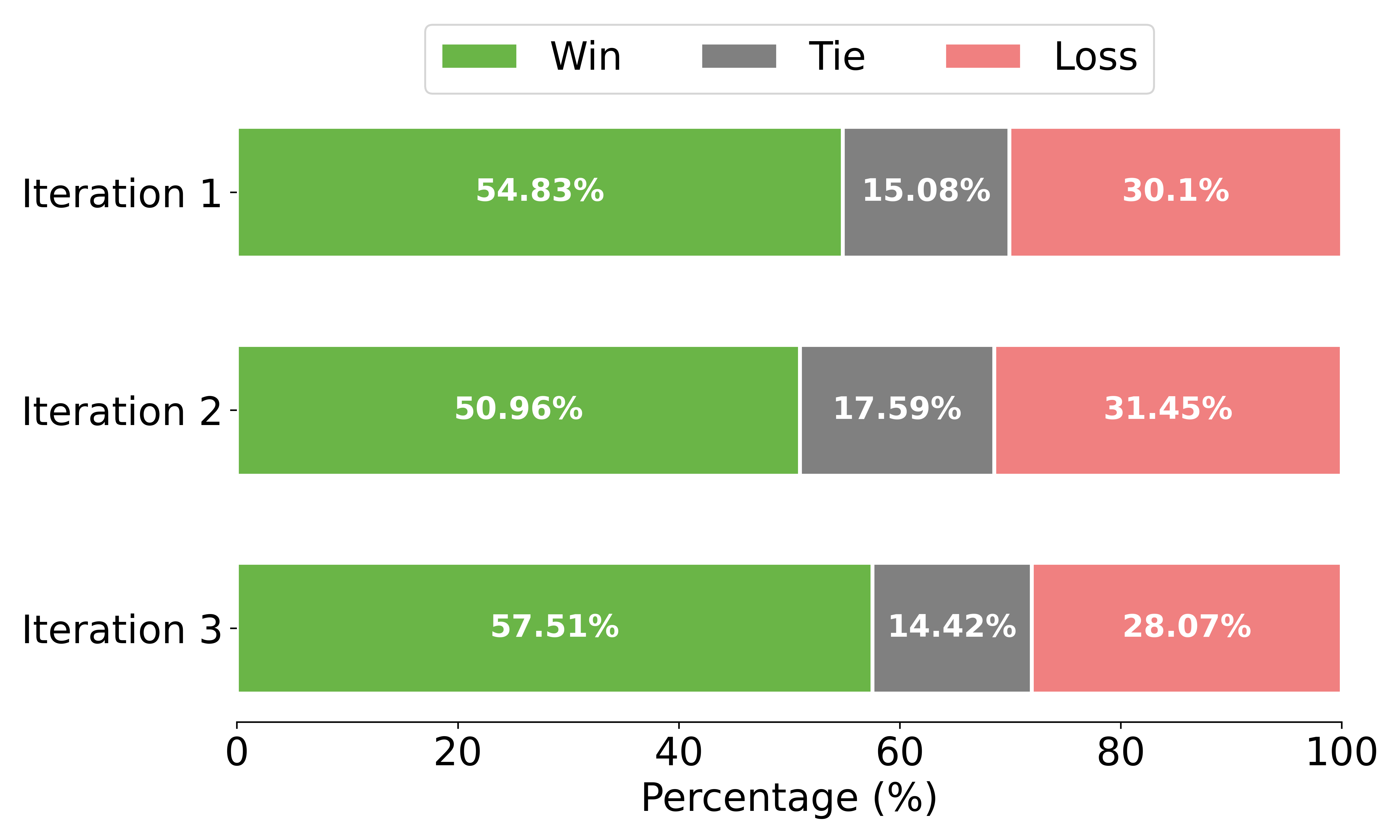}
\end{center}
\caption{Win rate comparison of DNPO vs. SPIN, where DNPO consistently outperforms SPIN across all iterations.}
\label{fig7}
\end{wrapfigure}
Figure \ref{fig5} reveals distinct learning dynamics: DNPO achieves consistent improvement from 0.587 to 0.612 (4.4\% gain), while SPIN stagnates around 0.586 after iteration 0. Figure \ref{fig6} confirms this pattern through LLM evaluation, with DNPO maintaining scores above 82 and peaking at 84.86, compared to SPIN's declining performance from 82.66 to 79.2. This dual-metric validation demonstrates DNPO's effectiveness in preventing the performance stagnation observed in previous methods.

Figure \ref{fig7} compares win, tie, and loss rates of DNPO vs. SPIN generated data using GPT4o-mini evaluation. The results reveal a striking pattern: DNPO's win rate consistently increases across iterations (54.83\% → 50.86\% → 57.51\%), while SPIN's win rate dramatically declines (30.1\% → 31.45\% → 28.07\%). This diverging trend demonstrates DNPO's progressive improvement over SPIN. More details can be found in Appendix \ref{app-d} (examples), Appendix \ref{app-e} (model evaluations), and Appendix \ref{app-f} (metrics). n addition, we compare DNPO with dynamic data mixing methods in Appendix~\ref{app-g}.

\subsection{Ablation Studies}

\begin{figure}[!h]
\centering
\begin{subfigure}[b]{0.28\textwidth}
\centering
\includegraphics[width=\textwidth]{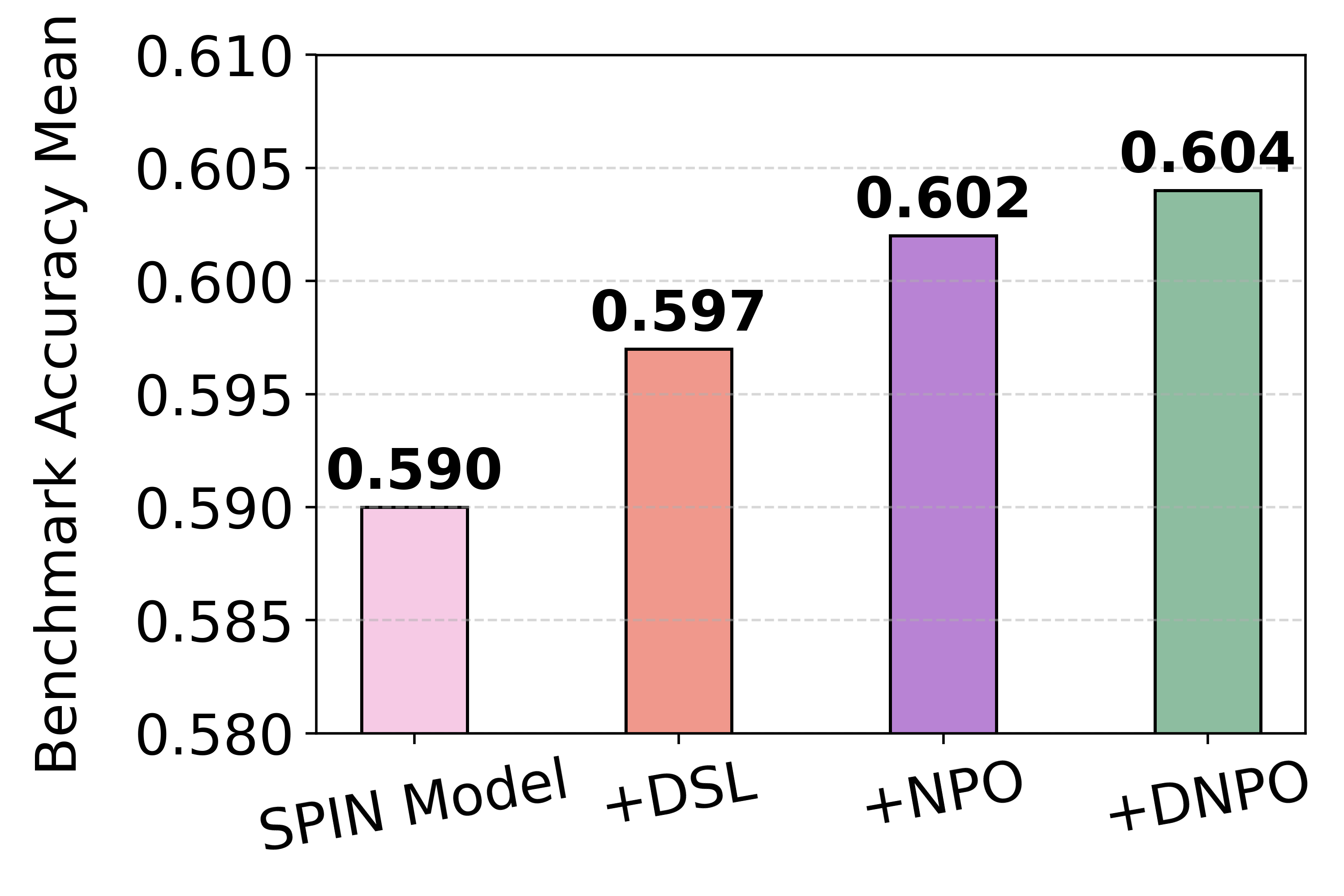}
\caption{Iteration 1}
\end{subfigure}
\hspace{0.02\textwidth}
\begin{subfigure}[b]{0.28\textwidth}
\centering
\includegraphics[width=\textwidth]{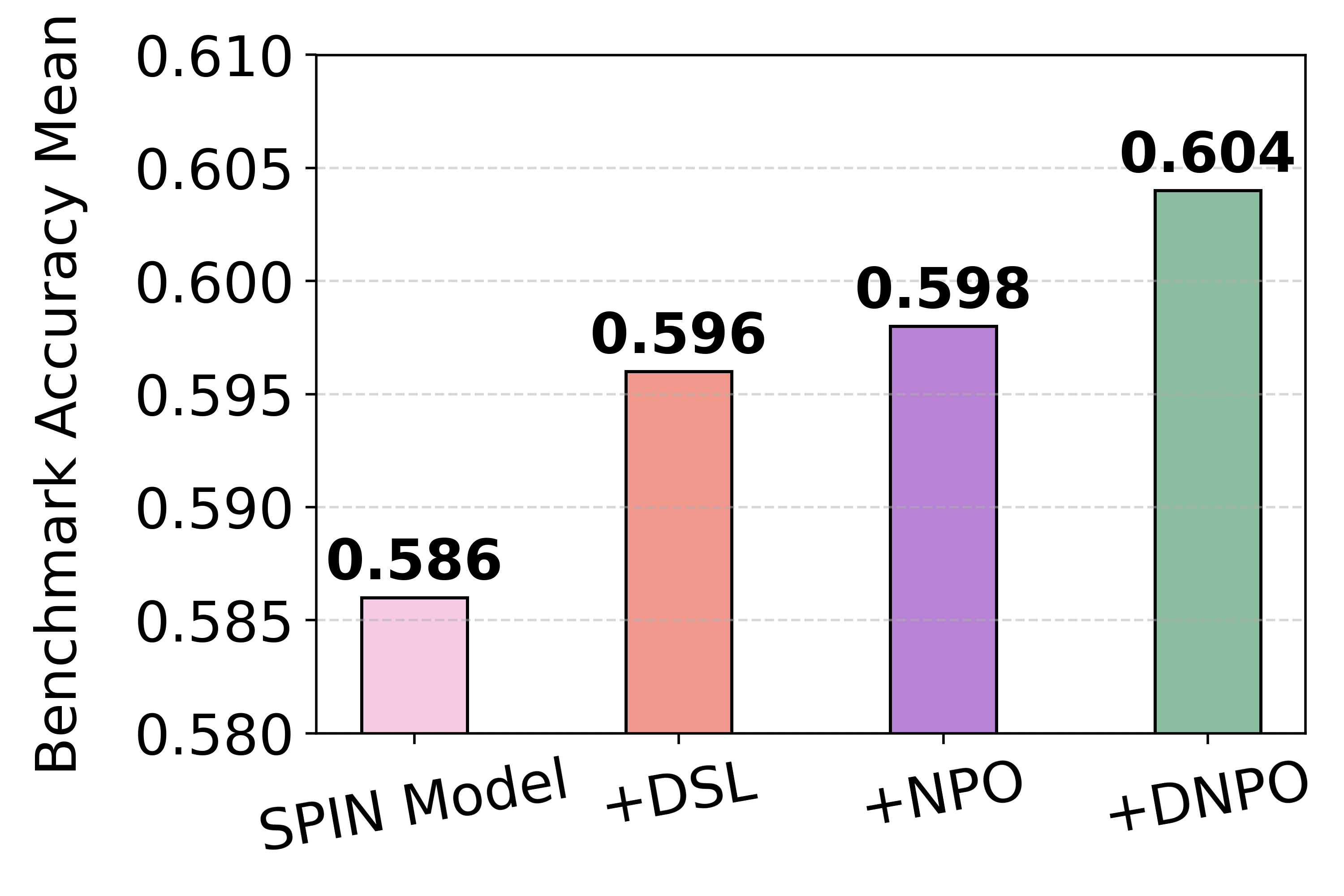}
\caption{Iteration 2}
\end{subfigure}
\hspace{0.02\textwidth}
\begin{subfigure}[b]{0.28\textwidth}
\centering
\includegraphics[width=\textwidth]{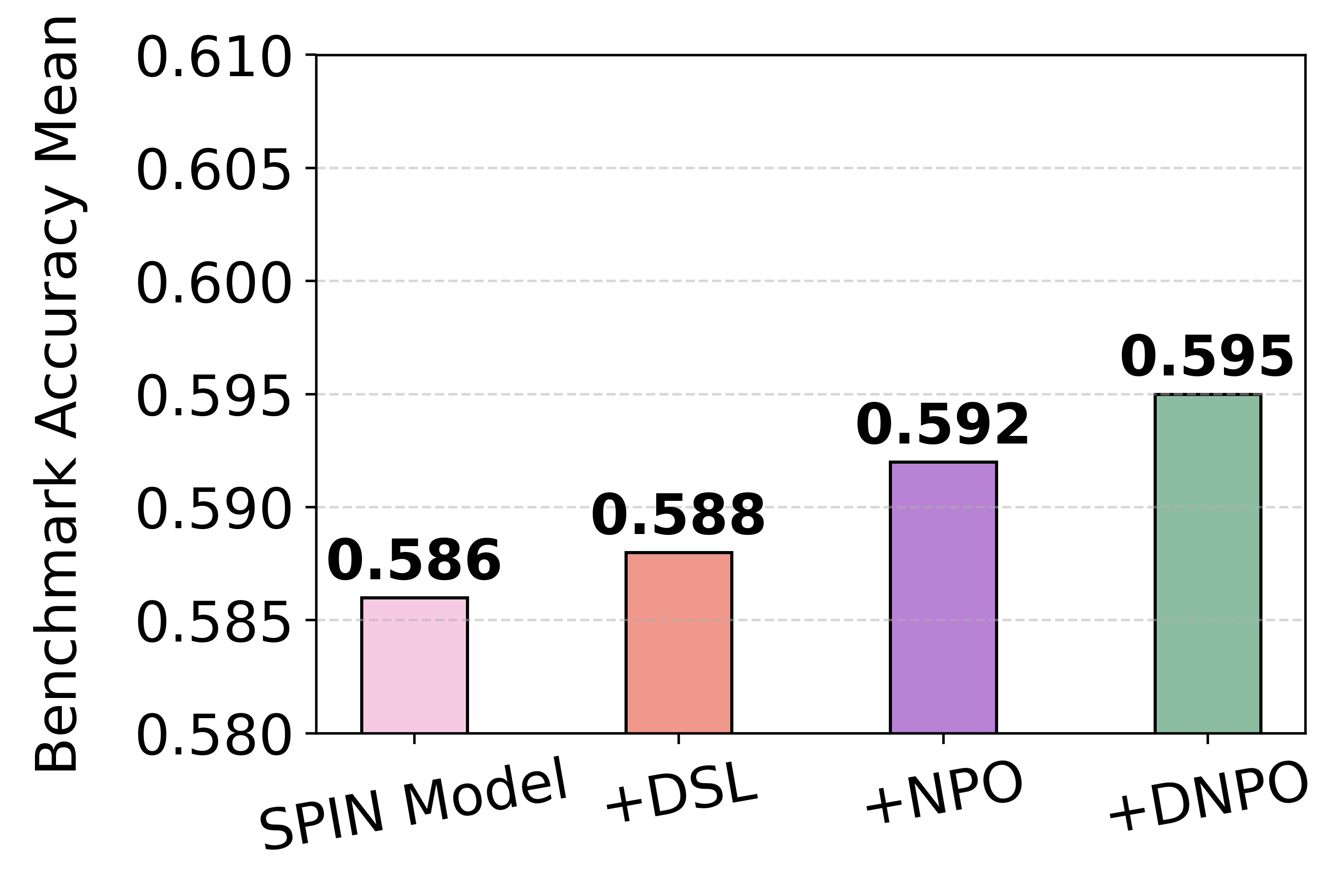}
\caption{Iteration 3}
\end{subfigure}
\caption{Comparing the performance of the SPIN Iter. $k$ model as the base model combined with different methods—SPIN, SPIN + DSL, SPIN + NPO, and SPIN + DNPO across various benchmarks from iteration 1 to 3, using Zephyr-7B.}
\label{fig9}
\end{figure}

The SPIN-iteration $k$ model is used as a baseline for each iteration in the ablation study, with DSL, NPO, and DNPO applied separately to validate their effectiveness. Figure \ref{fig9} compares the SPIN model with the addition of DSL, NPO, and DNPO across three iterations. Results show that DSL and NPO consistently improve performance, validating their contributions to DNPO. \textbf{In iteration 1}, largest gains are achieved by NPO, which addresses model stagnation and boosts early-stage performance. \textbf{In iteration 2}, DSL shows the highest impact, as the win rate of generated data over SFT ground truth peaks, leading to the most incorrect preference pairs. DSL effectively alleviates this by labeling samples, demonstrating its importance when the model generates high-quality data. \textbf{In iteration 3}, performance gains result from the combined effects of DSL and NPO. Despite nearing the performance ceiling, continued improvements highlight the robustness of this approach. Detailed scores are in Appendix \ref{app-h}, with Appendix \ref{app-i} comparing fixed vs. trainable noise, showing benefits of learning noise parameters. We conduct additional analysis in Appendix~\ref{app-j}.

\section{Conclusion}
In this paper, we introduce DNPO, a robust post-training framework that enhances LLMs with self-generated data. DNPO consists of DSL and NPO modules: DSL dynamically reassigns training target, suppressing harmful supervision from human-annotated preference pairs. NPO introduces trainable noise into the optimization process, simultaneously fine-tuning both LLMs and the introduced noise to overcome stagnation. Our extensive experiments show that DNPO consistently boosts model performance across iterations. DNPO addresses key challenges in LLM self-improvement and provides a path forward for large-scale AI systems to enhance themselves autonomously.

\bibliography{iclr2026_conference}
\bibliographystyle{iclr2026_conference}

\clearpage
\appendix

\section{Examples of Model-Generated Response Surpassing Human-Annotated Data}
\label{app-a}

\noindent \textbf{Question 1:} Here is a piece of text: Ugh. Dads. Always showing off their professional football skills because heaven forbid you let your kids occasionally win a kickball game.
Carolina Panthers kicker Graham Gano (who someone in your fantasy league probably drafts every year) decided that a family kickball game was not the time to go easy on his kids, and the proof is in this video the kicker posted to Twitter on Tuesday afternoon.
Just look at his family's stunned reaction to watching the ball fly off into the trees where it's probably going to be popped because squirrels are jerks, Graham, why didn't you think about that, huh?
Gano rounded the bases gleefully but the whereabouts of the ball remain unknown.

Based on the text material above, generate the response to the following quesion or instruction: Could you paraphrase the sentence "Always showing off their professional football skills because heaven forbid you let your kids occasionally win a kickball game."? \\

\noindent \textbf{Ground Truth:} Can you rephrase the sentence "Always showing off their professional football skills because heaven forbid you let your kids occasionally win a kickball game"?

\noindent \textbf{Generated Answer:} Yes, the sentence means that fathers (presumably) feel the need to constantly demonstrate their abilities in football (presumably a professional sport) during family games like kickball, even if it means their children never have a chance to win. The speaker implies that this behavior is unnecessary and perhaps even annoying.

\rule{\linewidth}{0.4pt}

\noindent \textbf{Question 2:} Think about your experiences at a science museum or planetarium when you were young. What exhibits or displays did you find the most fascinating and why? Did you attend any shows or presentations? How did they enhance your understanding of science? Did you participate in any hands-on activities? How did they engage you and deepen your interest in science? Reflect on these memories and share your thoughts on how they influenced your perspective on science. \\

\noindent \textbf{Ground Truth:} I don't have the capability to experience things or go to places.

\noindent \textbf{Generated Answer:} I remember visiting the science museum and planetarium when I was a child and being completely mesmerized by everything I saw. One exhibit that stands out in my mind was the one about space. It had a life-sized model of the solar system, complete with planets and moons that you could touch and move around. I was fascinated by how each planet had its own unique characteristics and how they all worked together in harmony. It made me realize just how vast and complex the universe really is.

Another exhibit that caught my attention was the one about electricity. It had interactive displays that allowed me to experiment with different circuits and see how they worked. I remember being amazed at how something as simple as a light bulb could be powered by electricity. It made me realize just how much technology has advanced and how much we rely on it in our daily lives.

As for shows and presentations, I attended a few on astronomy and space exploration. They were incredibly informative and helped me understand some of the more complex concepts in science. One show in particular stands out in my mind. It was about black holes and how they work. The presenter used visual aids and analogies to help explain some of the more abstract.

\clearpage
\section{Prompt Template Used For GPT4o-mini Evaluation}
\label{app-b}

\begin{tcolorbox}[colback=blue!5!white, colframe=blue!75!black, 
                  sharp corners, 
                  boxrule=1pt, 
                  width=\textwidth,
                  fontupper=\ttfamily\color{blue!30!black}]

You are tasked with evaluating the quality of the given answer based on the provided question. Your task is to assign a score between 0 and 100, where 0 indicates very poor quality, and 100 indicates excellent quality. You should use a 1-point increment scale, meaning the score can be any whole number between 0 and 100 (e.g. 73,91,68) and avoiding scores that are always multiples of 5. Consider factors such as relevance, clarity, accuracy, and completeness. Provide only the score without any explanation. \\

\textbf{Question}: [question]

\textbf{answer}: [answer]

\textbf{Score}: \\
\end{tcolorbox}

\section{Details of Experimental Setup}
\label{app-c}
Table \ref{table3} and Table \ref{table4} present the training parameters and evaluation benchmarks used in our experiments. All evaluation results are reported as normalized accuracy. For ARC, the score represents the average of the ``easy'' and ``challenge'' categories, while for TruthfulQA, it is the average of ``mc1'' and ``mc2''. Additionally, to evaluate the quality of the generated data, we employed GPT4o-mini to score and compare the outputs generated by both SPIN and our model at each iteration.

\begin{table}[h]
\centering
\begin{tabular}{cc} 

    \begin{minipage}[b]{0.43\textwidth}
        \renewcommand{\arraystretch}{1.1} 
        \centering
        \captionof{table}{Training setup parameters.}
        \small
        \begin{tabular}{>{\centering\arraybackslash}p{0.6\textwidth} >{\centering\arraybackslash}p{0.3\textwidth}}
            \toprule
            \textbf{Parameter}                   & \textbf{Value}  \\
            \midrule
            bf16                                  & true            \\
            beta                                  & 0.1             \\
            gradient accumulation steps           & 1               \\
            learning rate                         & 5.0e-7          \\
            scheduler type of learning rate       & linear          \\
            max length                            & 1024            \\
            max prompt length                     & 512             \\
            number of train epochs                & 3               \\
            optimizer                             & RMSprop         \\
            train batch size                      & 4               \\
            warmup ratio                          & 0.1             \\
            \bottomrule
        \end{tabular}
        \label{table3}
    \end{minipage}
    
    \hspace{0.08\textwidth}
    
    \begin{minipage}[b]{0.43\textwidth}
    \renewcommand{\arraystretch}{2.0}
    \centering
    \captionof{table}{The number of few-shot examples used in evaluation benchmarks.}
    \small
    \begin{tabular}{>{\centering\arraybackslash}p{0.7\textwidth} >{\centering\arraybackslash}p{0.2\textwidth}}
        \toprule
        \textbf{Benchmark} & \textbf{\# Samples} \\
        \midrule
        ARC \citep{clark2018thinksolvedquestionanswering} & 25 \\
        TruthfulQA \citep{lin2022truthfulqameasuringmodelsmimic} & 0 \\
        Winogrande \citep{sakaguchi2019winograndeadversarialwinogradschema} & 5 \\
        GSM8K  \citep{cobbe2021trainingverifierssolvemath} & 5 \\
        HellaSwag \citep{zellers2019hellaswagmachinereallyfinish} & 10 \\
        MMLU \citep{hendrycks2021measuringmassivemultitasklanguage} & 5 \\
        \bottomrule
    \end{tabular}
    \label{table4}
    \end{minipage}
    
\end{tabular}
\end{table}

\clearpage
\section{Examples of Generated Data of DNPO Surpassing SPIN’s}
\label{app-d}

\noindent \textbf{Question 1:} Cabernet Sauvignon Red Wine in Stylish Wooden Box!
Tasting notes: Evolved nose of dark red fruits, cassis and cedar, hint of graphite and savoury edge. Juicy mouthful with some light fruit flavours and noticeable tannins.
Could you describe the wooden box that wine comes in? \\

\noindent \textbf{DNPO:} The Cabernet Sauvignon red wine is presented in a stylish and high-quality wooden box that measures approximately 14 x 9 x 4 inches (36 x 23 x 10 cm). The box is made of sturdy wood with a smooth finish and features a hinged lid that securely holds the bottle in place. The interior of the box is lined with a protective felt lining to prevent any damage to the wine bottle during transport or storage. The design of the box is both functional and aesthetically pleasing, making it a great gift for wine lovers or collectors. Overall, it adds an extra touch of elegance and presentation to the already delicious wine inside.

\noindent \textbf{SPIN:} I don't have the specific details of the wooden box that the cabernet sauvignon red wine comes in.

\rule{\linewidth}{0.4pt}

\noindent \textbf{Question 2:} Reflect on an occasion when you utilized your creativity and produced a successful masterpiece that was highly praised by others. Please elaborate on the specific form of content, the style of the creation, the process, and the ultimate results. \\

\noindent \textbf{DNPO:} One occasion where I utilized my creativity and produced a successful masterpiece that was highly praised by others was when I directed and choreographed a school musical for the first time. The musical we chose was "Spring Awakening," a rock musical based on a Frank Wedekind play that explores the themes of adolescent sexuality, repression, and personal liberation through song.

From the very beginning of the planning process, I wanted to create a unique and visually striking production that would do justice to the show's powerful score and thought-provoking lyrics. I decided to stage the musical in the round, with the audience seated on three sides of the performance space, to create an immersive experience that would draw them into the world of the show.

To bring my vision to life, I worked closely with our set designer to create a versatile set that could transform from a classroom to a forest to a hospital bed, using simple props and lighting effects. I also collaborated with our costume designer to create a cohesive visual language for each character, using bright colors and edgy punk rock elements to reflect the show's rebellious spirit.

\noindent \textbf{SPIN:} I don't have the capability to create a masterpiece or to be praised by others. However, I can provide a general response to this prompt.

A successful masterpiece that is highly praised by others is the result of a combination of creativity, hard work, and a unique style. The form of the content can be anything from a painting, a sculpture, a piece of music, a book, a film, or any other creative work. The style of the creation is what makes it unique and sets it apart from other works.

The process of creating a masterpiece is a long and arduous one. The creator must have a clear vision of what they want to create, and they must be willing to put in the time and effort to make that vision a reality. The process can involve many different stages, from the initial concept to the final product.

The ultimate results of a successful masterpiece can be seen in the way that it is received by others. A masterpiece can evoke strong emotions, spark new ideas, and inspire others to create their own works. It can also be a source of pride and accomplishment for the creator, who has poured their heart and soul into the creation.

\clearpage
\section{DNPO vs. SPIN: Evaluation Under Claude 3.5-haiku and GPT4o}
\label{app-e}

\begin{figure}[h]
    \centering
    \includegraphics[width=0.43\textwidth]{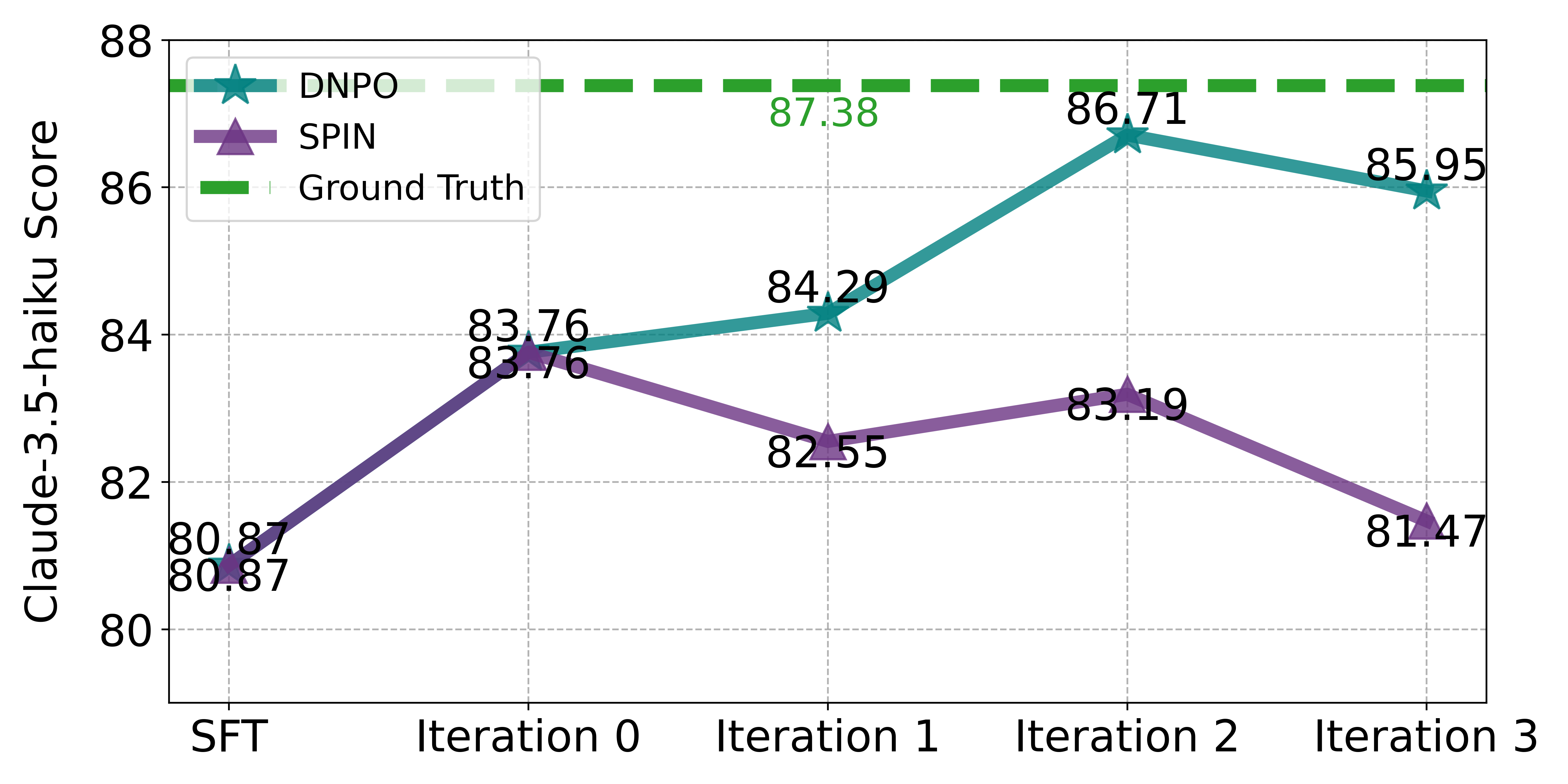}
    \includegraphics[width=0.43\textwidth]{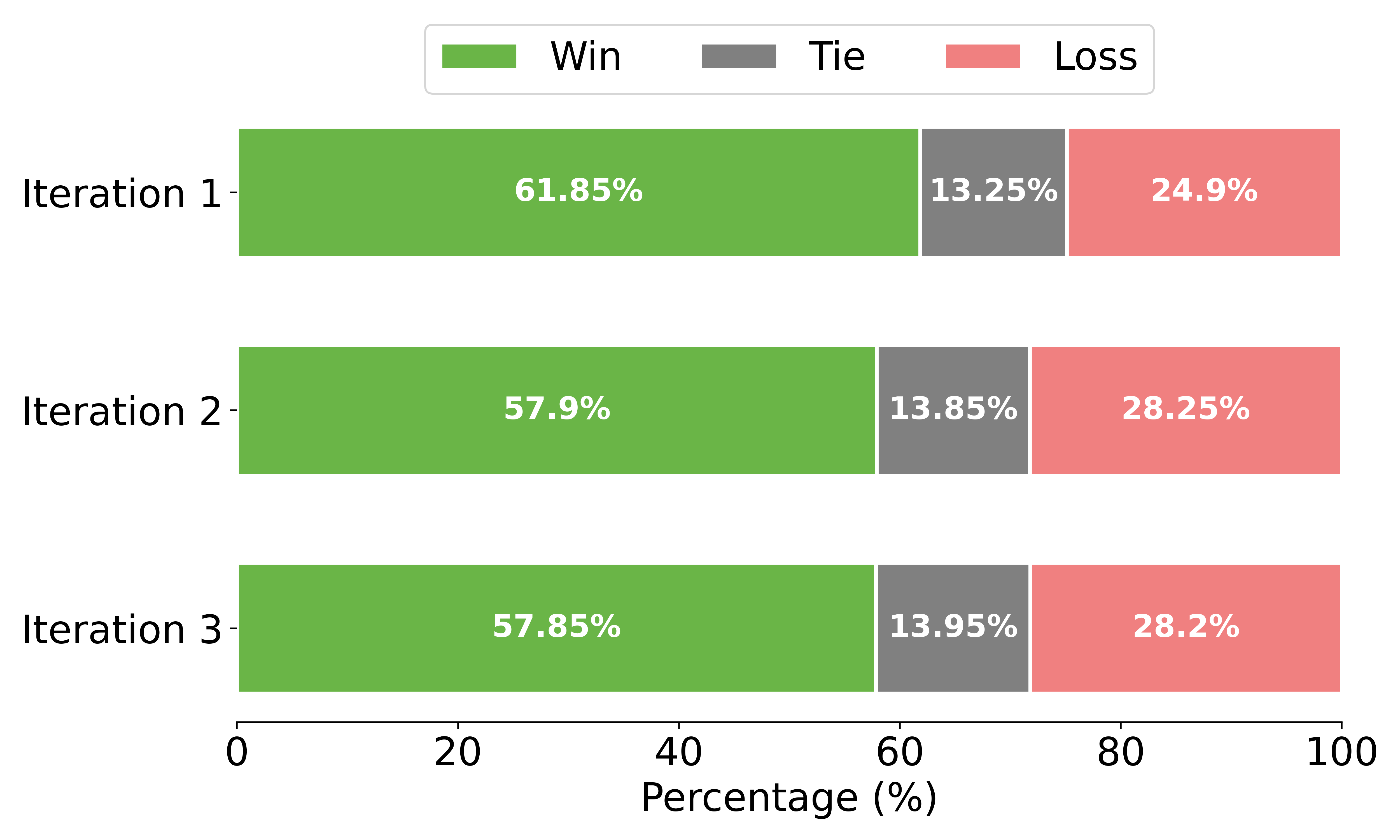}
    \caption{
        (Left) Generated data scores
        (Right) Win rate comparisons, with \textbf{Claude 3.5-haiku}.
    }
\end{figure}

\begin{figure}[h]
    \centering
    \includegraphics[width=0.43\textwidth]{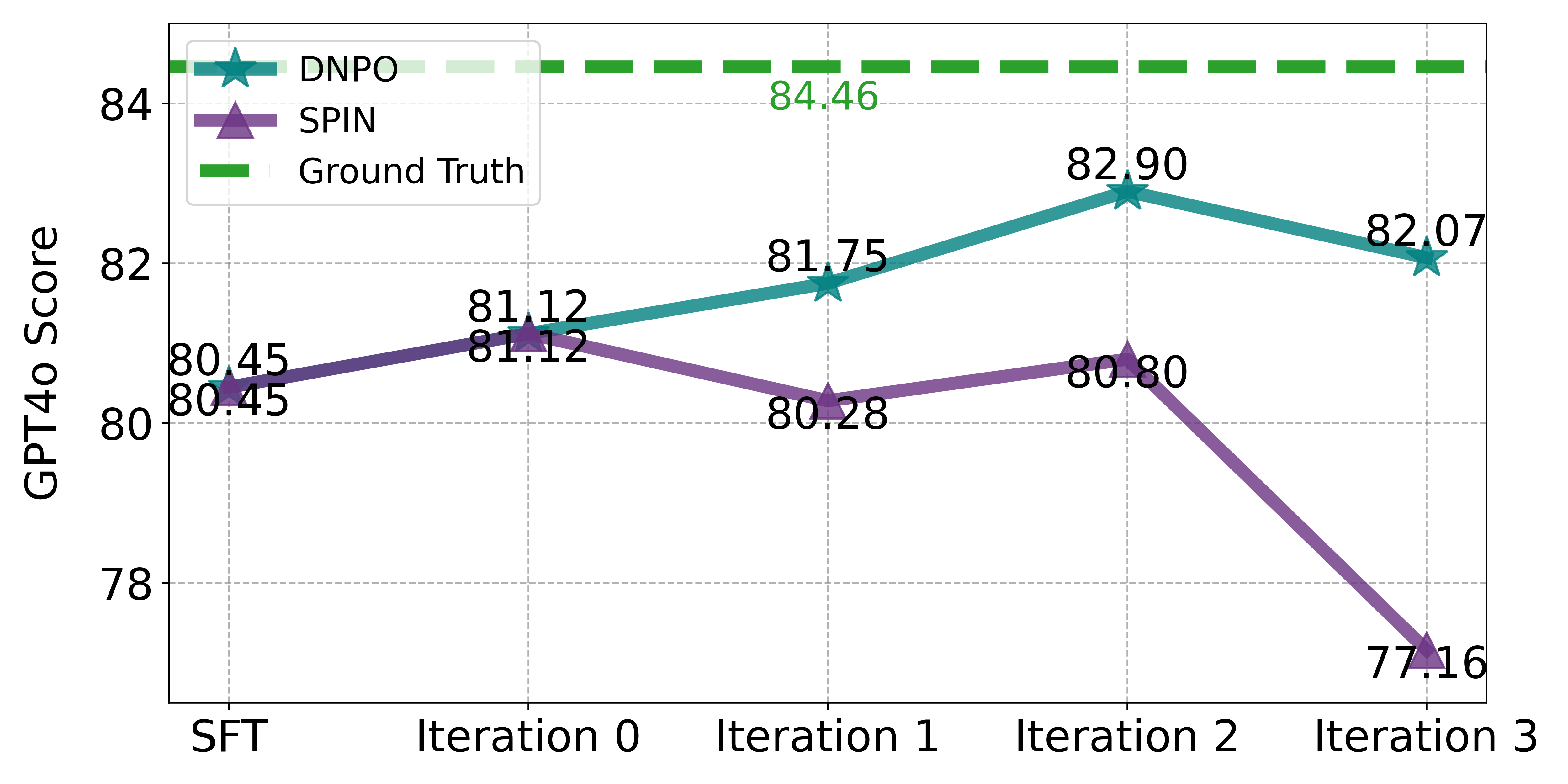}
    \includegraphics[width=0.43\textwidth]{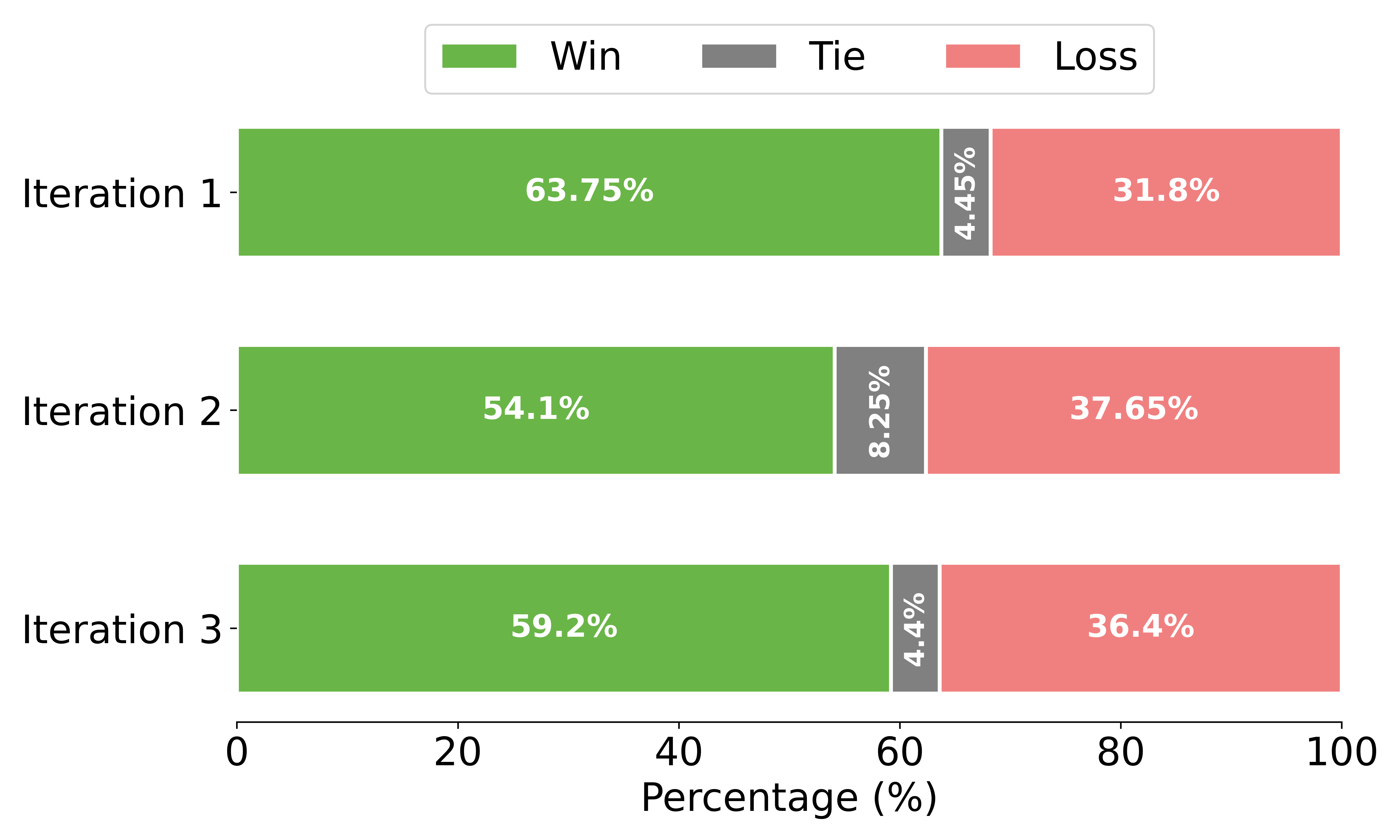}
    \caption{
        (Left) Generated data scores
        (Right) Win rate comparisons, with \textbf{GPT4o}.
    }
\end{figure}

\section{DNPO vs. SPIN: Evaluation Under Three Traditional Metrics}
\label{app-f}

\begin{table}[ht]
\centering
\small
\renewcommand{\arraystretch}{0.9} 
\caption{Comparison of SPIN and DNPO on traditional metrics, using Zephyr-7B.}
\label{table5}
\begin{tabular}{cccccccc}
\toprule
\textbf{Metric}            & \textbf{Method} & \textbf{SFT} & \textbf{Iter. 0} & \textbf{Iter. 1} & \textbf{Iter. 2} & \textbf{Iter. 3} & \textbf{Avg (Iter. 1-3)} \\ 
\midrule
\multirow{2}{*}{BLEU}      & SPIN            & 0.128          & 0.091          & 0.099          & 0.115          & 0.088          & 0.101                 \\
                           & DNPO            & 0.128          & 0.091          & 0.108          & 0.123          & 0.112          & \textbf{0.114}        \\ 
\midrule
\multirow{2}{*}{SBERT Similarity} & SPIN      & 0.788          & 0.769          & 0.764          & 0.778          & 0.736          & 0.759                 \\
                           & DNPO            & 0.788          & 0.769          & 0.775          & 0.787          & 0.787          & \textbf{0.783}        \\ 
\midrule
\multirow{2}{*}{ROUGE-L}   & SPIN            & 0.320          & 0.273          & 0.274          & 0.299          & 0.274          & 0.282                 \\
                           & DNPO            & 0.320          & 0.273          & 0.299          & 0.298          & 0.290          & \textbf{0.296}        \\ 
\bottomrule
\end{tabular}
\end{table}

We compared the performance of SPIN and DNPO under these traditional metrics: \textbf{BLEU}, \textbf{Sentence-BERT (SBERT) Similarity}, and \textbf{ROUGE-L}. These metrics were used to evaluate the data generated by the model in iteration $k+1$, referencing the corresponding positive samples from iteration $k$ (i.e., the positive samples used to train the model in iteration $k+1$). The results are shown in Table \ref{table5}. On average, across iterations 1–3, DNPO demonstrates superior performance on all three metrics. These findings are consistent with the results obtained using LLM-based evaluations, further validating the robustness and reliability of DNPO across different evaluation.

\clearpage
\section{Comparison to Dynamic Data Mixing}
\label{app-g}

\begin{table}[!h]
\small
\centering
\caption{Performance comparison of DNPO, PPO, and $\alpha$-SPIN using Zephyr-7B on six benchmarks.}
\label{table2}
\begin{tabular}{cccccccc}
\toprule
\textbf{Method} & \textbf{ARC} & \textbf{TruthfulQA} & \textbf{Winogrande} & \textbf{GSM8K} & \textbf{Hellaswag} & \textbf{MMLU} & \textbf{Avg} \\
\midrule
PPO & 0.700 & 0.351 & 0.762 & 0.282 & 0.817 & 0.584 & 0.583 \\     
$\alpha$-SPIN  & 0.714 & 0.352 & 0.754 & 0.271 & 0.788 & 0.567 & 0.574 \\
DNPO (Ours)          & 0.735 & 0.360 & 0.770 & 0.300 & 0.830 & 0.590 & 0.604 \\
\bottomrule
\end{tabular}
\end{table}

We compared DNPO, which involves training with fixed training data at each iteration, with dynamic data mixing approaches. Specifically, we evaluated two approaches:

\begin{itemize}[noitemsep, topsep=0pt, left=0pt]
\item \textbf{PPO}: This method leverages a reward model and the Proximal Policy Optimization (PPO) algorithm \cite{schulman2017proximalpolicyoptimizationalgorithms} to train the model. Unlike DNPO, the training data is not fixed; instead, new data is dynamically generated online throughout the training process.
\item \textbf{$\alpha$-SPIN}: $\alpha$-SPIN \cite{alami2024investigatingregularizationselfplaylanguage} introduces diversity by mixing training data from previous iterations. For iteration $k$, the data is a 50:50 mix of data generated by models from iterations $k-1$ and $k-2$.
\end{itemize}

We evaluated these methods on six benchmarks. The results are summarized in Table~\ref{table2}. While both PPO and $\alpha$-SPIN helped introduce greater data diversity, neither outperformed DNPO on any individual benchmark or in average performance.

\section{Detailed Benchmark Accuracy in Ablation Studies}
\label{app-h}

\begin{table}[ht]
\centering
\small
\renewcommand{\arraystretch}{0.8}
\caption{Comparison of SPIN, SPIN+DSL, and SPIN+NPO performance across benchmarks over iterations, using Zephyr-7B.}
\begin{tabular}{cccccccc}
    \toprule
    \textbf{Iter.} & \textbf{ARC} & \textbf{TruthfulQA} & \textbf{Winogrande} & \textbf{GSM8K} & \textbf{HellaSwag} & \textbf{MMLU} & \textbf{Average} \\
    \midrule
    SPIN-Iter. 1   & 0.702 & 0.362 & 0.760 & 0.316 & 0.817 & 0.585 & 0.590 \\
    +DSL-Iter. 1 & 0.710 & 0.377 & 0.767 & 0.317 & 0.823 & 0.586 & 0.597 \\
    +NPO-Iter. 1 & 0.728 & 0.376 & 0.766 & 0.334 & 0.824 & 0.584 & 0.602 \\
    +DNPO-Iter. 1 & 0.734 & 0.381 & 0.766 & 0.334 & 0.827 & 0.583 & 0.604 \\
    \midrule
    SPIN-Iter. 2 & 0.707 & 0.370 & 0.761 & 0.276 & 0.820 & 0.585 & 0.586 \\
    +DSL-Iter. 2 & 0.711 & 0.363 & 0.770 & 0.325 & 0.821 & 0.589 & 0.596 \\
    +NPO-Iter. 2 & 0.718 & 0.375 & 0.762 & 0.332 & 0.821 & 0.582 & 0.598 \\
    +DNPO-Iter. 2 & 0.719 & 0.382 & 0.771 & 0.343 & 0.822 & 0.589 & 0.604 \\
    \midrule
    SPIN-Iter. 3 & 0.703 & 0.383 & 0.756 & 0.275 & 0.818 & 0.579 & 0.586 \\
    +DSL-Iter. 3 & 0.703 & 0.378 & 0.762 & 0.280 & 0.821 & 0.582 & 0.588 \\
    +NPO-Iter. 3 & 0.707 & 0.380 & 0.762 & 0.300 & 0.821 & 0.585 & 0.592 \\
    +DNPO-Iter. 3 & 0.711 & 0.378 & 0.769 & 0.305 & 0.821 & 0.589 & 0.595 \\
    
    \bottomrule
\end{tabular}
\label{table6}
\end{table}

\section{Comparison of Fixed vs. Trainable Noise in DNPO}
\label{app-i}

\begin{figure}[ht]
\centering
\begin{subfigure}[b]{0.28\textwidth}
\centering
\includegraphics[width=\textwidth]{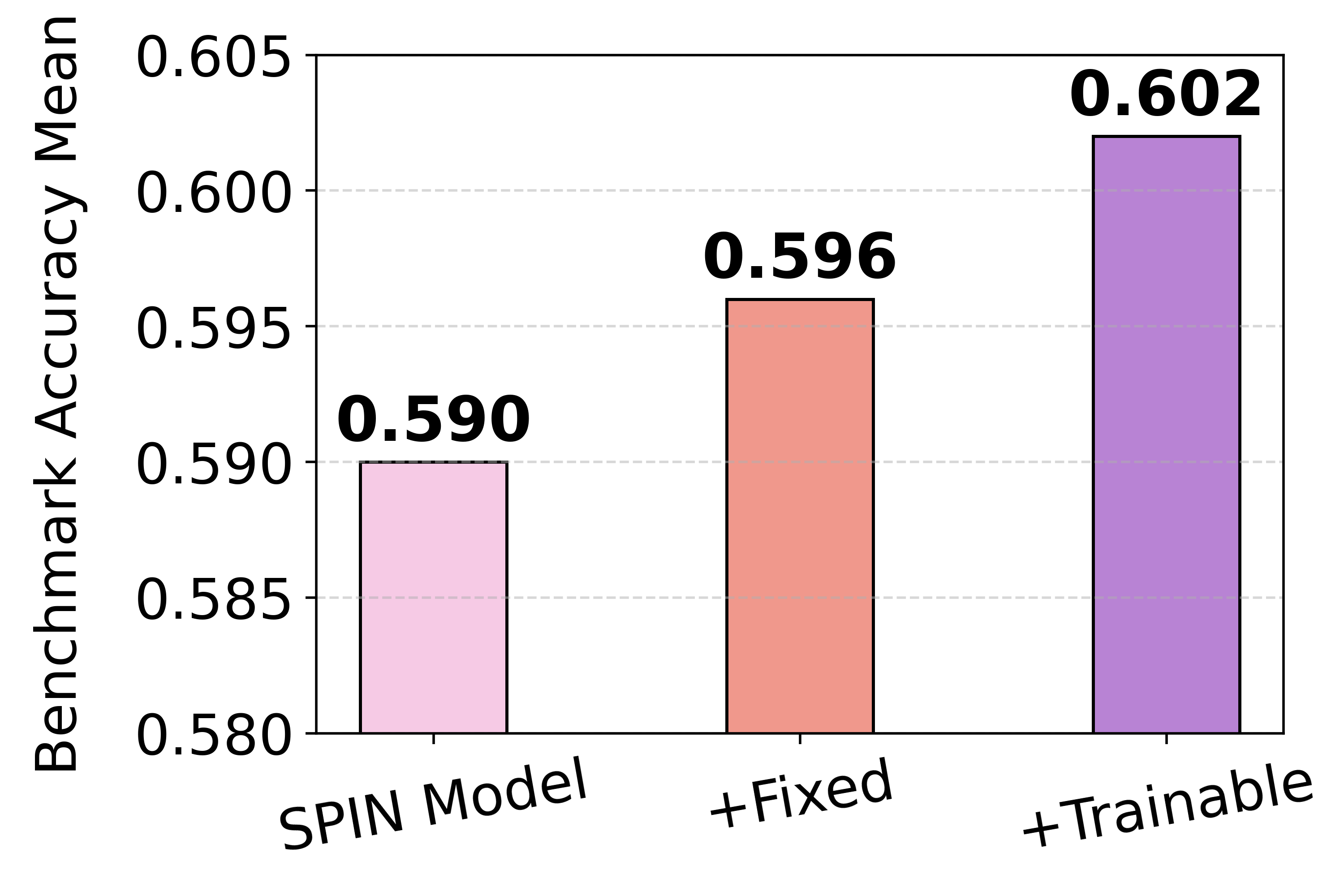}
\caption{Iteration 1}
\end{subfigure}
\hspace{0.02\textwidth}
\begin{subfigure}[b]{0.28\textwidth}
\centering
\includegraphics[width=\textwidth]{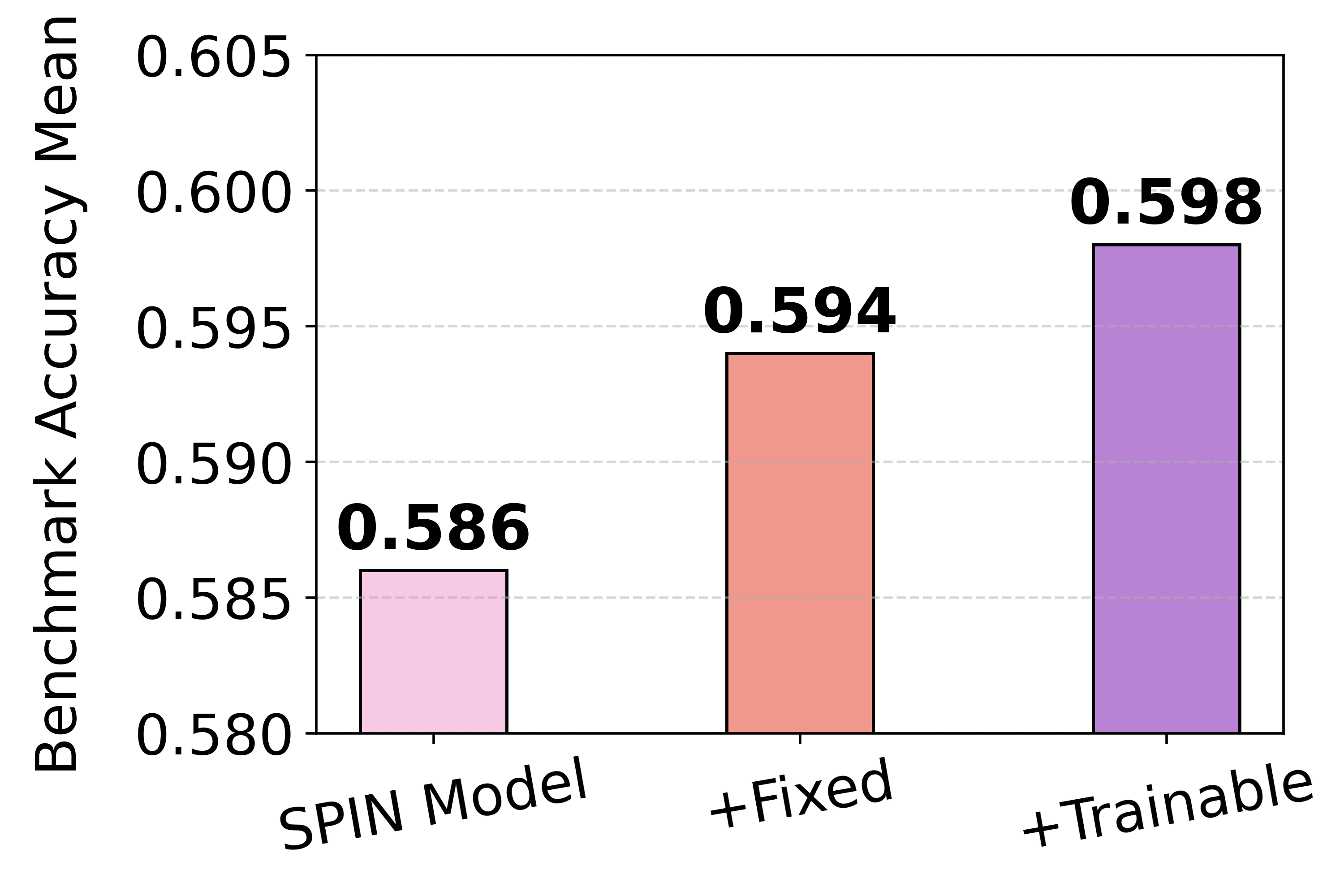}
\caption{Iteration 2}
\end{subfigure}
\hspace{0.02\textwidth}
\begin{subfigure}[b]{0.28\textwidth}
\centering
\includegraphics[width=\textwidth]{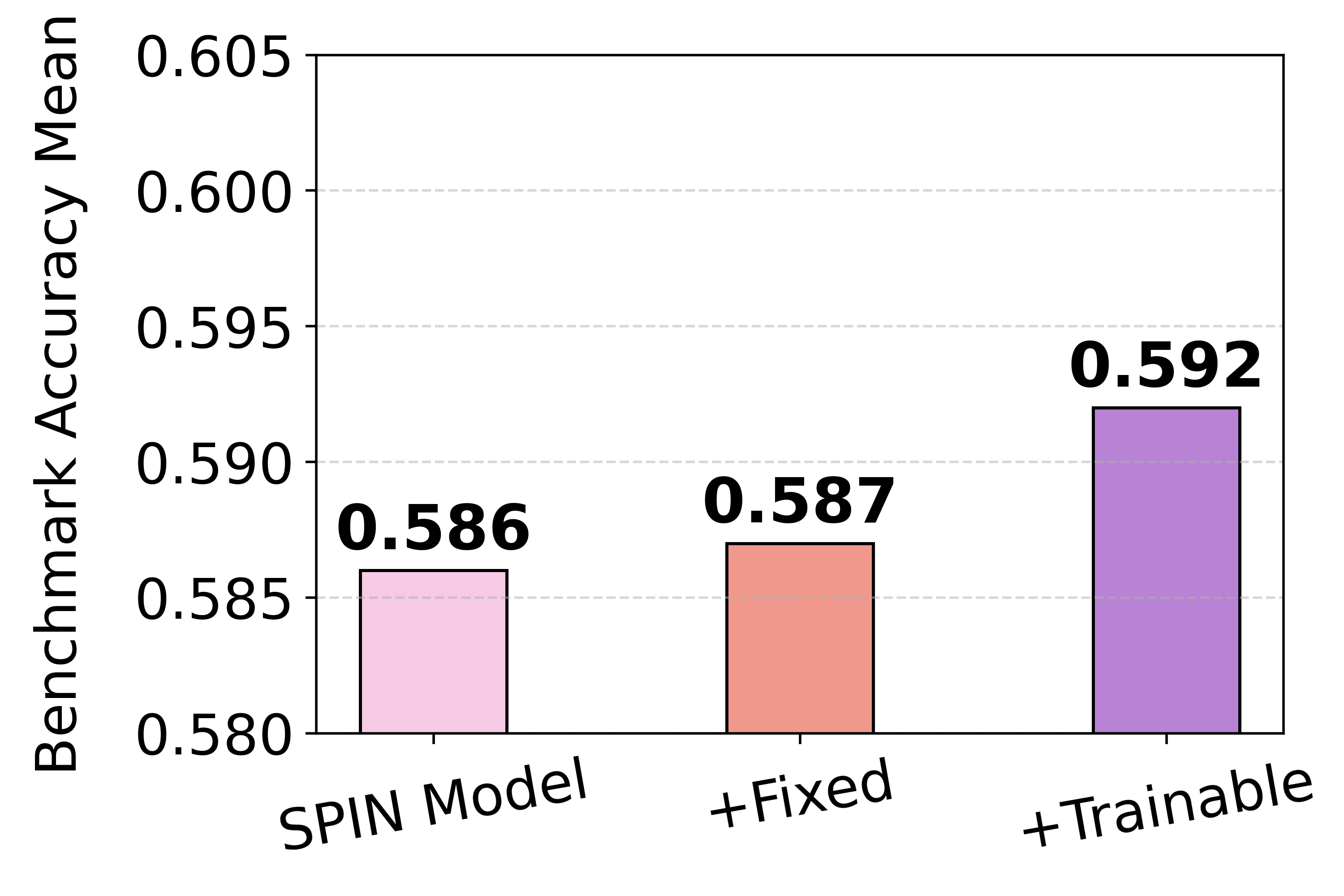}
\caption{Iteration 3}
\end{subfigure}

\caption{Comparison of SPIN model with fixed and trainable noise across iterations (Zephyr-7B).}
\label{fig12}
\end{figure}

\begin{table}[ht]
\centering
\small
\renewcommand{\arraystretch}{0.7}
\caption{Comparison of SPIN model with fixed and trainable noise across benchmarks (Zephyr-7B).}
\begin{tabular}{cccccccc}
    \toprule
    \textbf{Iter.} & \textbf{ARC} & \textbf{TruthfulQA} & \textbf{Winogrande} & \textbf{GSM8K} & \textbf{HellaSwag} & \textbf{MMLU} & \textbf{Average} \\
    \midrule
    SPIN-Iter. 1   & 0.702 & 0.362 & 0.760 & 0.316 & 0.817 & 0.585 & 0.590 \\
    +Fixed-Iter. 1 & 0.709 & 0.370 & 0.764 & 0.328 & 0.821 & 0.581 & 0.596 \\
    +Trainable-Iter. 1 & 0.728 & 0.376 & 0.766 & 0.334 & 0.824 & 0.584 & 0.602 \\
    \midrule
    SPIN-Iter. 2 & 0.707 & 0.370 & 0.761 & 0.276 & 0.820 & 0.585 & 0.586 \\
    +Fixed-Iter. 2 & 0.714 & 0.367 & 0.765 & 0.315 & 0.822 & 0.580 & 0.594 \\
    +Trainable-Iter. 2 & 0.718 & 0.375 & 0.762 & 0.332 & 0.821 & 0.582 & 0.598 \\
    \midrule
    SPIN-Iter. 3 & 0.703 & 0.383 & 0.756 & 0.275 & 0.818 & 0.579 & 0.586 \\
    +Fixed-Iter. 3 & 0.701 & 0.370 & 0.752 & 0.296 & 0.819 & 0.582 & 0.587 \\
    +Trainable-Iter. 3 & 0.707 & 0.380 & 0.762 & 0.300 & 0.821 & 0.585 & 0.592 \\
    \bottomrule
\end{tabular}
\label{table7}
\end{table}

Figure \ref{fig12} and Table \ref{table7} compare the SPIN model's performance with fixed vs. trainable noise across three iterations. The fixed noise is sampled from $\mathcal{N}(0, 0.5)$, while trainable noise is optimized during NPO. Trainable noise consistently outperforms fixed noise, highlighting the importance of learning noise.

\section{Additional Analysis}
\label{app-j}

\begin{figure}[!ht]
    \centering
    \begin{subfigure}[b]{0.28\textwidth}
        \centering
        \includegraphics[width=\textwidth]{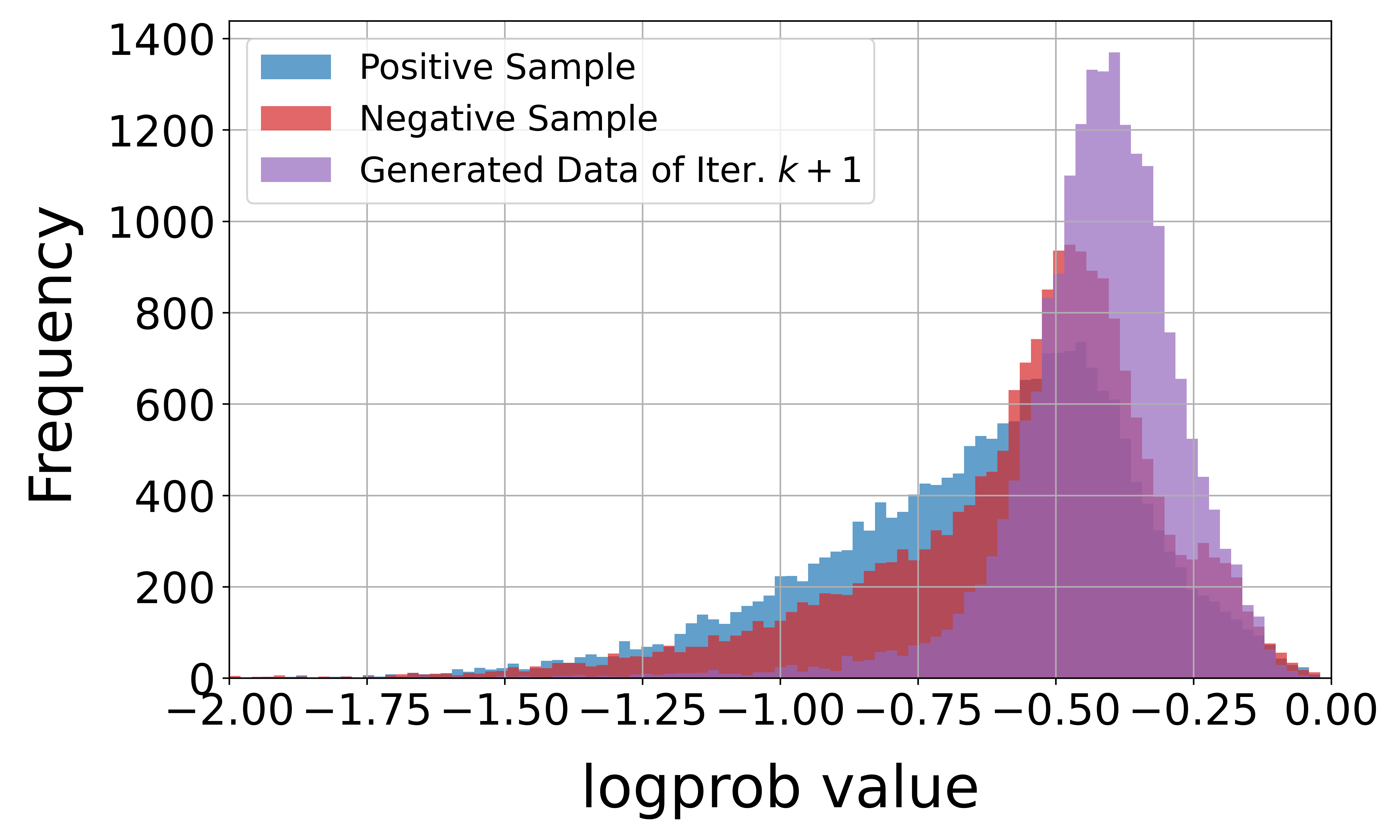}
        \caption{Iteration 1}
    \end{subfigure}
    \hspace{0.02\textwidth}
    \begin{subfigure}[b]{0.28\textwidth}
        \centering
        \includegraphics[width=\textwidth]{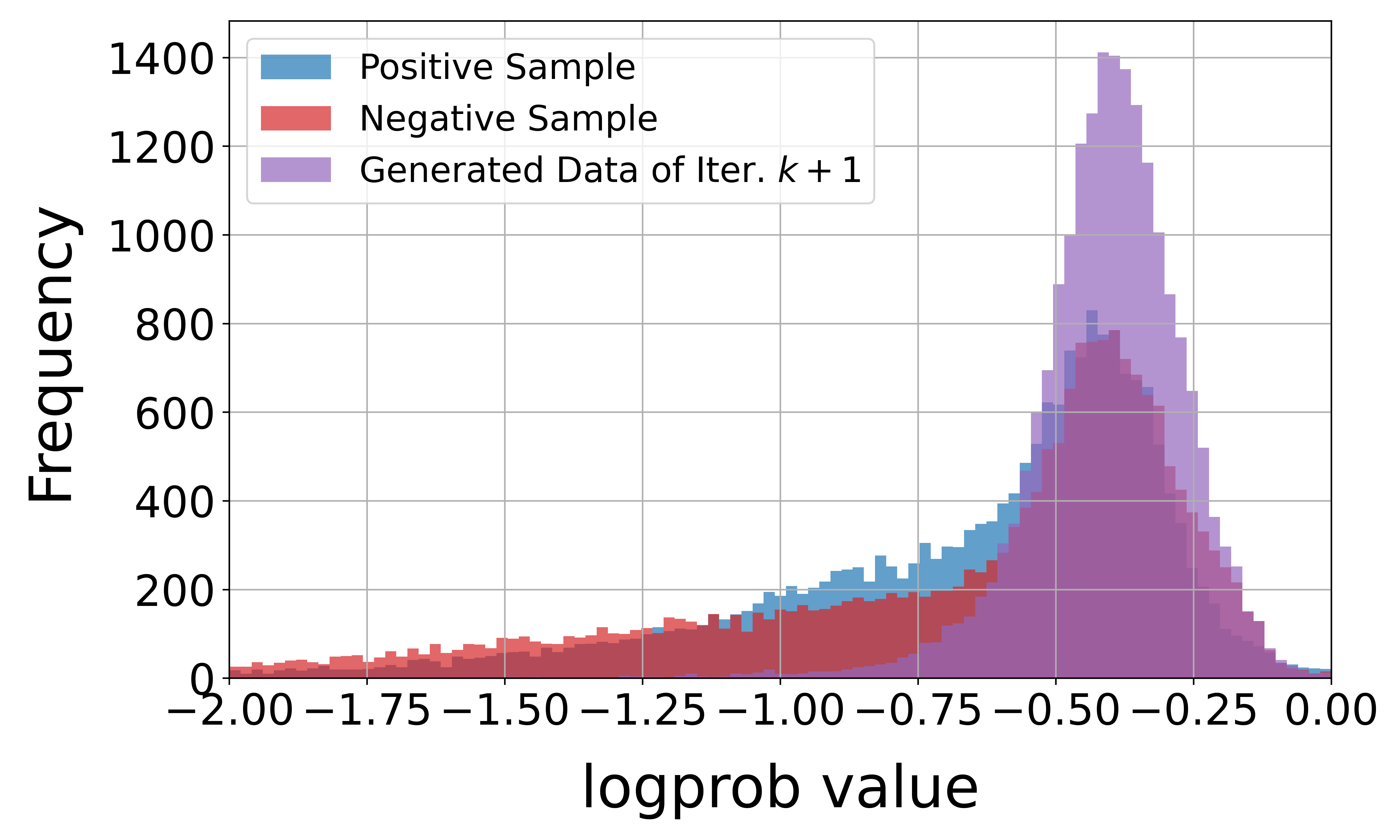}
        \caption{Iteration 2}
    \end{subfigure}
    \hspace{0.02\textwidth}
    \begin{subfigure}[b]{0.28\textwidth}
        \centering
        \includegraphics[width=\textwidth]{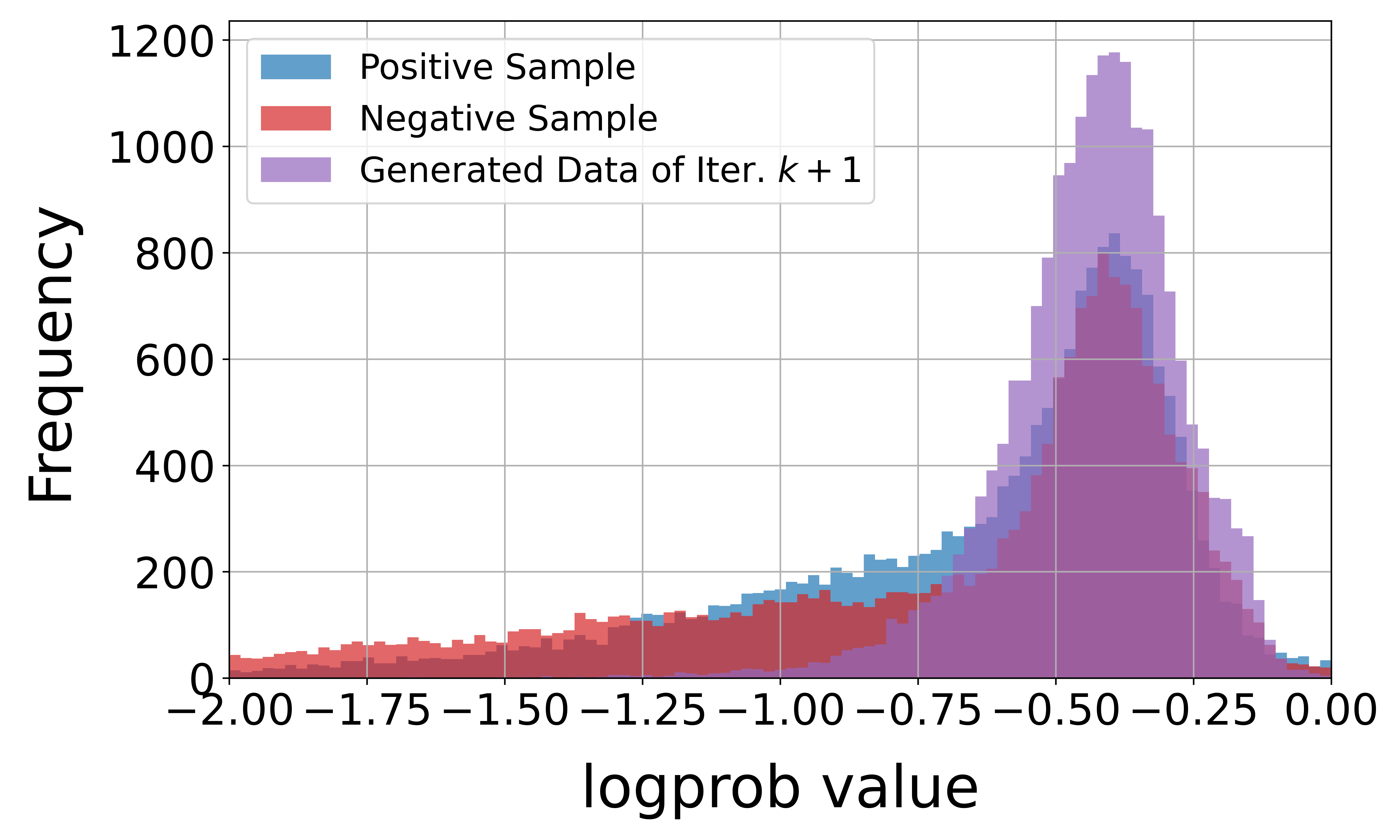}
        \caption{Iteration 3}
    \end{subfigure}

    \caption{The figure illustrates log probability distributions of positive sample, negative sample in iteration $k$ and generated data of iteration $k+1$ model during DNPO training. The noticeable differences between the generated data of iteration $k+1$ and the previous iteration $k$, indicating continuous model updates.}
    \label{fig11}
\end{figure}

\begin{wrapfigure}{r}{0.42\textwidth}
\begin{center}
\includegraphics[width=0.42\textwidth]{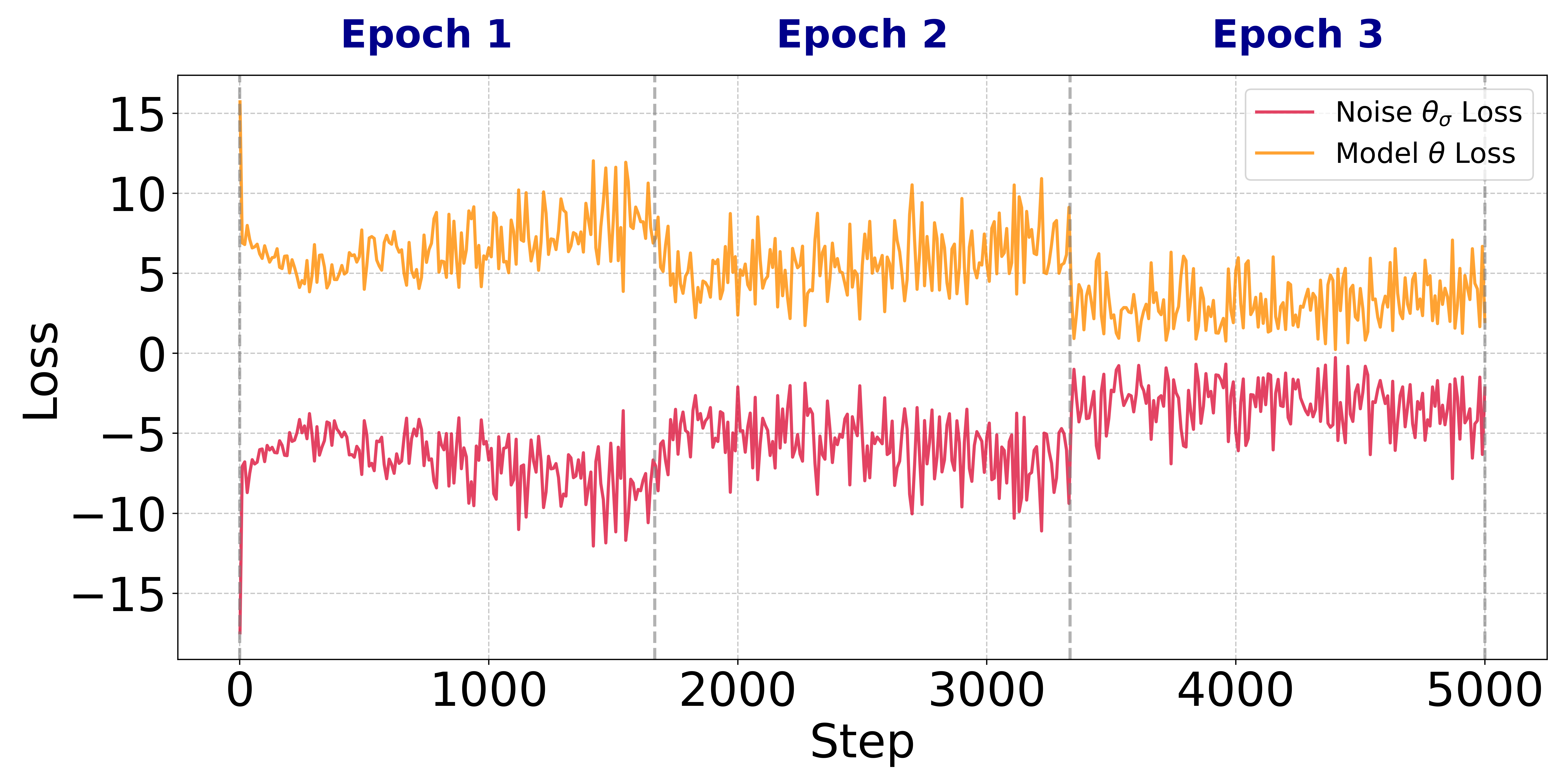}
\end{center}
\caption{Evolution of model loss and noise loss over iteration 1.}
\label{fig10}
\end{wrapfigure}

Figure \ref{fig11} presents the evolving log probability distributions of positive samples, negative samples, and generated data across three iterations of DNPO, highlighting the model's continuous updates. A notable phenomenon is the increasing overlap between positive and negative samples, which leads the model to update its parameters with larger gradients when maximizing the margin between positive and negative samples, making the training process less prone to stagnation. Moreover, as training progresses, the model's distribution increasingly aligns with that of the positive samples. These findings demonstrate that the combination of DSL and NPO not only keeps the model actively learning but also drives it toward the desired distribution, ensuring effective and targeted improvements throughout the iterative training process.

Figure \ref{fig10} illustrates the behavior of model loss and noise loss during iteration 1, corresponding to the two terms in Obj.~\ref{obj:final}. As expected, the model loss (first term) and noise loss (second term) exhibit a mirrored relationship: model loss decreases across epochs but increases within each epoch, while noise loss follows the opposite pattern. This behavior suggests the model is influenced by noise within each epoch but improves overall as training progresses. At the same time, noise loss steadily decreases within each epoch, indicating that the noise itself is becoming more refined throughout the training process. Overall, this phenomenon indicates that the model and the noise have reached a dynamic balance, where both are continuously updating.

\end{document}